%% file: _main.tex
\input{_constants}
\arxiv
\pdfoutput=1
\documentclass[10pt,twocolumn,letterpaper]{article}
\input{cvpr_header}
\begin{document}
\title{\paperTitle}
\author{\authorBlock}
\maketitle

\input{00_abstract}
\input{01_intro}

\input{02_related}
\input{03_method}
\input{04_experiments}

\input{10_conclusion}

{\small
\bibliographystyle{ieeenat_fullname}
\bibliography{11_references}
}

\ifarxiv \clearpage \appendix \input{12_appendix} \fi

\end{document}

%% file: _constants.tex
\def\paperID{14225}
\def\confName{CVPR}
\def\confYear{2024}

\def\paperTitle{MAFA: Managing False Negatives for Vision-Language Pre-training}

\def\authorBlock{
    Jaeseok Byun$^{1}$\thanks{Equal contribution} \qquad
    Dohoon Kim$^{1}$\footnotemark[1] \qquad
    Taesup Moon$^{1,2}$\thanks{Corresponding author} \\
    $^{1}$ Department of ECE, Seoul National University \\
    $^{2}$ Department of ASRI/INMC/IPAI/AIIS, Seoul National University \\
    {\tt\small \{wotjr3868, dohoon.kim, tsmoon\}@snu.ac.kr }
}

\newif\ifreview 
\newif\ifarxiv \newcommand{\arxiv}{\arxivtrue}
\newif\ifcamera 
\newif\ifrebuttal 

%% file: cvpr_header.tex
\ifreview \usepackage[review]{cvpr} \fi
\ifarxiv \usepackage[pagenumbers]{cvpr} \fi
\ifrebuttal \usepackage[rebuttal]{cvpr} \fi
\ifcamera \usepackage{cvpr} \fi

\input{_macros}  

\usepackage{xr-hyper}

\makeatletter
\newcommand*{\addFileDependency}[1]{
  \typeout{(#1)}
  \@addtofilelist{#1}
  \IfFileExists{#1}{}{\typeout{No file #1.}}
}

\makeatother

\usepackage{graphicx}
\usepackage{multirow}
\usepackage{pifont} 
\newcommand{\cmark}{\ding{51}}%
\newcommand{\xmark}{\ding{55}}%
 
\usepackage{bm}

\usepackage{xcolor}
\usepackage[ruled, linesnumbered]{algorithm2e}

\definecolor{cvprblue}{rgb}{0.21,0.49,0.74}
\usepackage[pagebackref,breaklinks,colorlinks,citecolor=cvprblue]{hyperref}
\usepackage[capitalize]{cleveref}
\crefname{section}{Sec.}{Secs.}
\crefname{table}{Table}{Tables}
\crefname{figure}{Fig.}{Figs.}

%% file: _macros.tex
\usepackage{graphicx}	
\usepackage{amsmath}	
\usepackage{amssymb}	
\usepackage{booktabs}
\usepackage{times}
\usepackage{microtype}
\usepackage{epsfig}
\usepackage[table,xcdraw,dvipsnames]{xcolor}
\usepackage{caption}
\usepackage{float}
\usepackage{placeins}
\usepackage{color, colortbl}
\usepackage{stfloats}
\usepackage{enumitem}
\usepackage{tabularx}
\usepackage{xstring}
\usepackage{multirow}
\usepackage{xspace}
\usepackage{url}
\usepackage{subcaption}
\usepackage{xcolor}
\usepackage[hang,flushmargin]{footmisc}

\ifcamera \usepackage[accsupp]{axessibility} \fi

% ...

\ifarxiv  \fi

\newcommand{\R}[1]{{%
    \textbf{%
        \ifstrequal{#1}{1}{\textcolor{red}{R#1}}{%
        \ifstrequal{#1}{2}{\textcolor{blue}{R#1}}{%
        \ifstrequal{#1}{3}{\textcolor{magenta}{R#1}}{%
        \ifstrequal{#1}{4}{\textcolor{teal}{R#1}}{%
                           \textcolor{cyan}{R#1}%
        }}}}%
    }%
}}

%% file: 00_abstract.tex
\begin{abstract}


 We consider a critical issue of false negatives in Vision-Language Pre-training (VLP), a challenge that arises from the inherent \textit{many-to-many correspondence} of image-text pairs in large-scale web-crawled datasets. 
 The presence of false negatives can impede achieving optimal performance and even lead to a significant performance drop.
To address this challenge, we propose MAFA (\textbf{MA}naging \textbf{FA}lse negatives), which consists of two pivotal components building upon the recently developed GRouped mIni-baTch sampling (GRIT) strategy: 1) an efficient connection mining process that identifies and converts false negatives into positives, 
and 2) \textit{label smoothing} for the image-text contrastive (ITC) loss. 
Our comprehensive experiments verify the effectiveness of MAFA across multiple downstream tasks, emphasizing the crucial role of addressing false negatives in VLP, potentially even surpassing the importance of addressing false positives.
In addition, the compatibility of MAFA with the recent BLIP-family model is also demonstrated. Code is available at \href{https://github.com/jaeseokbyun/MAFA}{https://github.com/jaeseokbyun/MAFA}.

\end{abstract}

%% file: 01_intro.tex
\section{Introduction}
\label{sec:intro}

With large-scale web-crawled datasets \cite{(CC)sharma2018conceptual, changpinyo2021conceptual, (Laion400m)schuhmann2021laion, (Laion5b)schuhmann2022laion}, majorities of vision-language pre-training (VLP) models are trained in a self-supervised learning manner using the combination of several pre-tasks and losses \cite{(ALBEF)li2021align, (BLIP)li2022blip,(GRIT-VLP)byun2022grit,(TCL)yang2022vision, (X-VLM)eng2021multi}: \textit{e.g., }masked language modeling (MLM), image-text contrastive (ITC), and image-text matching (ITM) losses. 
Despite their promising results, one of the pressing challenges they face is the presence of \textit{noisy} captions in image-text pairs that often provide incomplete or even incorrect descriptions \cite{(DiHT)radenovic2023filtering,(rebuttal-5)maini2023t,(rebuttal-6)gadre2024datacomp,(rebuttal-7)wang2023too,(rebuttal-8)nguyen2024improving,(rebuttal-9)yuksekgonul2022and, (ECCVCaption)chun2022eccv}.
Several recent works have focused on addressing such issue of noisy correspondence in image-text pairs \cite{(NCR)huang2021learning,(MRL)hu2021learning,(SGRAF)diao2021similarity,(NLIP)huang2022nlip,(DiHT)radenovic2023filtering, (BLIP)li2022blip}. 
Notably, BLIP \cite{(BLIP)li2022blip} introduced a caption refinement process by leveraging an image captioning model and a filter to generate synthetic clean captions and remove noisy captions. Such process can be seen as correcting the \textit{false positives} that were injected by the noisy captions.


\begin{figure}[t]
   \centering
   \includegraphics[width=\linewidth]{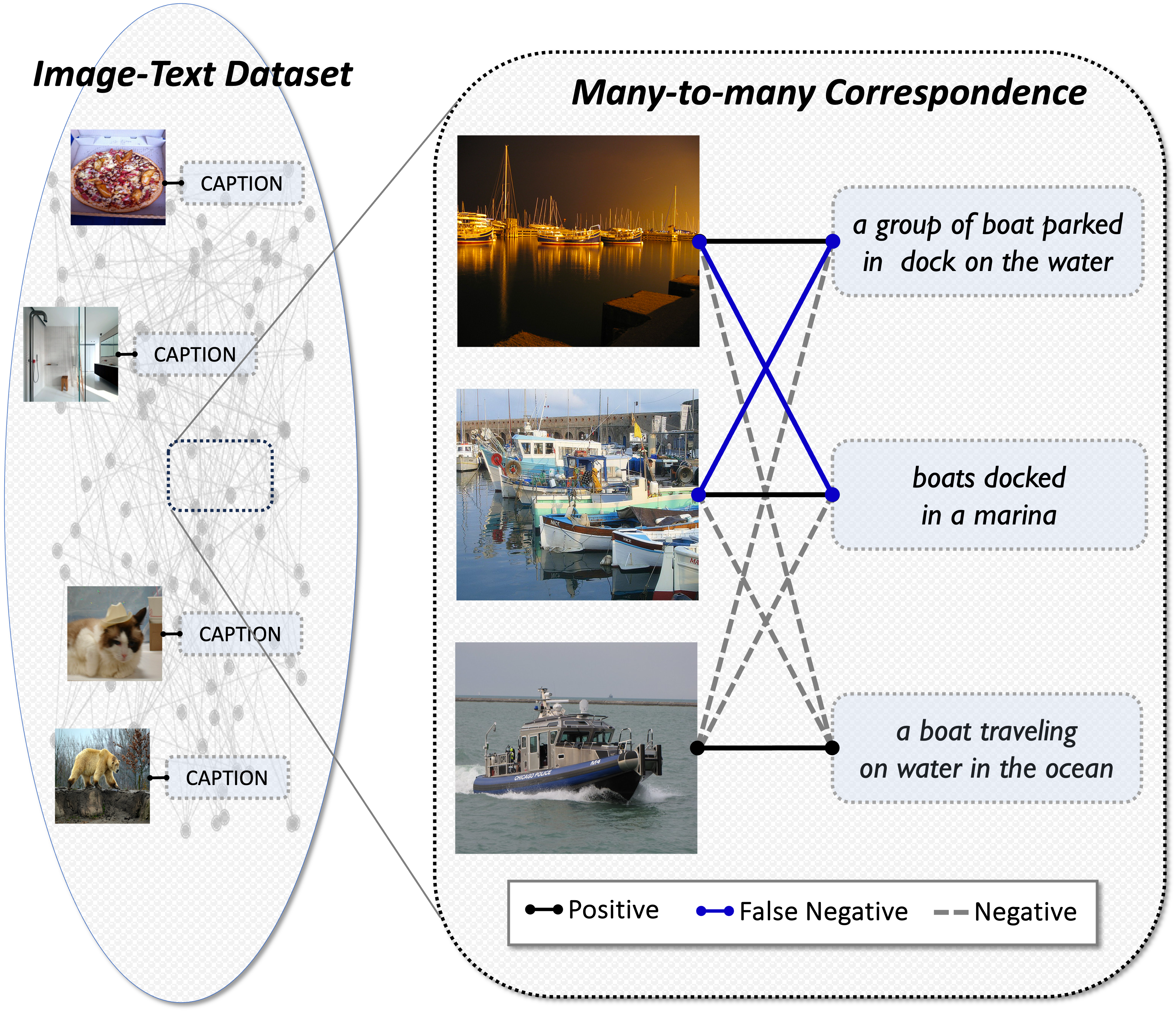}
   \vspace{-.2in}
    \caption{Examples of positives, negatives, and false negatives among image-text pairs. }
   \label{fig:falsenegatives}
   \vspace{-.1in}
\end{figure}

Contrastively, we note that there is another type of challenge for VLP that stems from the nature of many-to-many correspondence of image-text pairs. 
Namely, it is common for an image (resp. text) to have \textit{additional} positive connections (blue lines in Figure \ref{fig:falsenegatives}) with other texts (resp. images), which are paired with their corresponding images (resp. texts).
This is due to the fact that the existing image-text datasets are constructed by only collecting paired image-text instances, hence the information regarding non-paired but semantically close image-text combinations can be missed.
Consequently, for each image (resp. text), the text (resp. image) that is given as the pair with the image (resp. text) is treated as the only positive sample during pre-training, while the other texts (resp. images) are all treated as negatives. 
This setup inevitably leads to the prevalence of \textit{false negatives} during computing ITC and ITM losses and confuses the learning process. 
A naive solution would be to identify missing positive connections by examining all possible image-text combinations in the dataset. 
However, it is cleaerly infeasible for both manual and model-based evaluations due to prohibitive complexity. 
For example, even for a dataset of moderate size, \textit{e.g., }containing 5 million image-text pairs, the number of combinations that need to be examined is ${5M \choose 2}$, which amounts to approximately $12$ trillion.

We note such issue of false negatives has been more or less overlooked in recent studies \cite{(BLIP)li2022blip, (TCL)yang2022vision, (GRIT-VLP)byun2022grit}, which incorporated the in-batch hard negative sampling for the ITM task as a standard tool for VLP.
This sampling technique, initially proposed in ALBEF \cite{(ALBEF)li2021align}, involves selecting hard negative samples from the mini-batch based on the image-text similarity scores computed from the ITC task. 
More recently, GRIT-VLP \cite{(GRIT-VLP)byun2022grit} significantly improved the performance by proposing an improved hard negative sampling by first grouping the similar image-text pairs together before forming a mini-batch.
Ideally, if all semantically close image-text pairs were correctly labeled, hard negative mining could effectively identify only informative hard negatives. 
However, in a typical VLP setting where such information is absent, the hard negatives in fact frequently become false negatives, 
resulting in sub-optimal model performance.

To address this challenge of false negatives, particularly prevalent when hard negative sampling is in action, we have implemented two significant enhancements. 
Firstly, we devise an Efficient Connection Mining (ECM) process that identifies missing positive connections between non-paired but semantically close images and texts. 
Rather than reviewing all possible combinations, ECM strategically extracts the plausible candidates which are selected as hard negatives. 
These candidates are inspected by a pre-trained discriminator, which determines their potential to be converted into positives.
The candidates identified as positives by the discriminator are then incorporated as additional positives for calculating ITC, ITM, and MLM losses during the training process.
Secondly, we introduce Smoothed ITC (S-ITC) which is based on the principle of label smoothing \cite{(LabelSmoothing)szegedy2016rethinking}. 
This approach is specifically designed to mitigate the over-penalization of false negative samples within grouped mini-batches, without incurring any additional memory or computational overhead.

Our experimental results demonstrate that the proposed method, dubbed as MAFA (\textbf{MA}naging \textbf{FA}lse negatives), can substantially improve the VLP performance. For example, a model trained with MAFA on a standard 4M dataset (\textit{i.e., }4M-Noisy) can almost achieve the performance of a baseline model trained on a much larger 14M dataset, without exploiting any additional information such as bounding boxes or object tags. 
Our systematic ablation analyses demonstrate that such performance enhancement primarily results from mitigating effect of false negatives. 
Another finding from our experiments is that converting false negatives into additional positives is more advantageous than merely eliminating them. 
Moreover, we also demonstrate that the impact of addressing the false negative issue is orthogonal to and may outweigh that of addressing the false positive issue in VLP, which is done by comparing and combining MAFA with the BLIP \cite{(BLIP)li2022blip} framework. 
Finally, we show MAFA is also compatible with recently proposed BLIP-2 \cite{(BLIP-2)li2023blip}, underscoring the generality of our method in VLP.

%% file: 02_related.tex
\section{Related Work}
\label{sec:related}

\noindent{\textbf{Vision-language pre-training (VLP).}} 
Initial VLP models \cite{(UNITER)chen2020uniter,(Uni-VL)zhou2020unified,(VL-BART)cho2021unifying,(OSCAR)li2020oscar,(ViLT)kim2021vilt, (VILLA)gan2020large, (PixelBERT)huang2020pixel, (VilBERT)lu2019vilbert, (VL-BERT)su2019vl, (12in1)lu202012, (SOHO)huang2021seeing, (SimVLM)wang2021simvlm} which utilized a single multi-modal encoder,  
primarily employed random negative sampling during the ITM task. 
Recently, ALBEF \cite{(ALBEF)li2021align} incorporated the ITC loss and in-batch hard negative sampling strategy for ITM by leveraging image-text contrastive similarity scores.
Subsequently, the in-batch hard negative sampling strategy for ITM became an implicit rule for the BLIP-family models \cite{(TCL)yang2022vision, (X-VLM)eng2021multi, (GRIT-VLP)byun2022grit, (BLIP)li2022blip, (BUS)jiang2023bus, (VL-match)bi2023vl, (BLIP-2)li2023blip, (D-BLIP-2)jian2023bootstrapping} which adopt both ITC and ITM as training objectives.
While the significance of hard negative sampling for the ITM task has been highlighted in GRIT-VLP \cite{(GRIT-VLP)byun2022grit}, limited attention has been given to addressing the issue of false negatives arising from the hard negative samples. 
Existing studies have primarily focused on tackling false negatives only in the context of contrastive learning \cite{(PCME)chun2021probabilistic,(PVSE)ong2019polysemous, (ALBEF)li2021align, (GRIT-VLP)byun2022grit} with particular emphasis on the vision domain \cite{(Boosting)huynh2022boosting, (Debiased_hard)chuang2020debiased, (False_neg_incremental)chen2021incremental, (hardfalse)robinson2020contrastive}. 
To that end, we highlight the need for effective strategies to address false negatives in VLP and demonstrate that false negatives can be managed. \\

\noindent{\textbf{Label smoothing.}}  Label smoothing \cite{(LabelSmoothing)szegedy2016rethinking} is a widely adopted technique for improving generalization in various classification tasks. It converts the one-hot target labels into soft labels by mixing them with uniform distribution. This simple technique has demonstrated its efficacy in both visual \cite{(whenLS)muller2019does, (SLS)li2020regularization, (ALS)krothapalli2020adaptive} and language domains \cite{(LORAS)ghoshal2021learning, (Adaptivesmoothing)lee2022adaptive}. 
Its benefits have also led to its incorporation as a supplementary technique to enhance the fine-tuning of image-text models like CLIP \cite{(CLIP)radford2021learning} for image classification tasks \cite{(rebuttal-1)ilharco2022patching,(rebuttal-2)goyal2023finetune,(rebuttal-3)wortsman2022robust}. 
However, the application of label smoothing within VLP and its ability to address false negatives have not been thoroughly explored.
Recently, some studies \cite{(ALBEF)li2021align, (GRIT-VLP)byun2022grit} have introduced model-generated soft labels in the VLP domain.
However, we show that such soft labels are insufficient for effectively addressing false negatives, justifying the need to incorporate label smoothing to the contrastive loss when the hard negative sampling is employed.

\subsection{GRIT-VLP \cite{(GRIT-VLP)byun2022grit}}
GRIT-VLP uses ITC, ITM with in-batch hard negative sampling, and MLM as the objectives proposed in ALBEF, except the utilization of the momentum encoder in the pre-training. However, it significantly extends ALBEF by implementing two key components as follows: \\
\noindent{\textbf{(a) GRouped mIni-baTch (GRIT) sampling}} aims to construct mini-batches containing highly similar example groups. This facilitates the selection of informative hard negative samples during in-batch hard negative sampling. To avoid excessive memory or computational overhead, the procedure of constructing grouped mini-batches for the next epoch is performed concurrently with the loss calculation at each epoch. 
For this, additional queues are used to collect and search for the most similar examples, one by one, in which the similarity is measured by the ITC scores. These queues serve as the search space and are significantly larger than the mini-batch size $B$. Thus, the size of the queue, denoted as search space $M$, controls the level of hardness in selecting hard negative samples.
\\
\noindent{\textbf{(b) ITC with consistency loss}} 
attempts to address the issue of over-penalization in ITC that arises when GRIT is combined.
Contrary to ALBEF, similar examples are gathered in the GRIT-sampled mini-batch.
Thus, when one-hot labels are used for ITC, they result in equal penalization of all negatives, and it has been observed that the representations of similar samples may unintentionally drift apart. 
To mitigate this, GRIT-VLP incorporates soft pseudo-targets generated from the same pre-trained model as a mean of regularization.

%% file: 03_method.tex
\section{Motivation} 
\label{sec:motivation}
In order to quantify the tendency of the number of false negatives in the ITM task, we report a quantitative analysis result in 
Table \ref{table:FN_statistics}.  
We estimated the number of false negative pairs during a single epoch while training the ITM task, employing two distinct mini-batch sampling strategies: random sampling and GRIT sampling. These strategies were evaluated on both the original 4M dataset (4M-Noisy) and the BLIP-generated clean dataset (4M-Clean).
Given the infeasibility of manually examining every negative pair to determine whether it is a false negative, we utilized a strong ITM model pre-trained on a large-scale 129M dataset from BLIP --- \textit{i.e., }a negative pair is regarded as \textit{false} negative if the strong ITM model predicts it is ``matched''. While the ITM model does not always classify false negatives with perfect accuracy, its reliability is deemed adequate for approximating the trend in false negative counts. More details and analyses regarding counting the number of false negatives can be found in the Supplementary Material (S.M).

\begin{table}[t!]
\caption{Estimated number of false negatives (FN) for random sampling and GRIT sampling. 
The FNs are counted for each anchor image and text separately. 
The ratio (\%) represents the estimated proportion of FNs with respect to the total number of negative pairs used in ITM during a single epoch. We set the batch size $B$ as 96 for both samplings and $M=4800$ for GRIT sampling.
}
\vspace{-.1in}
\resizebox{\columnwidth}{!}{%
\begin{tabular}{c|c||c|c}
\hline
\textbf{Dataset}      & \textbf{Sampling} & \textbf{FN w.r.t. image} & \textbf{FN w.r.t. text} \\ \hline
\multirow{2}{*}{4M-Noisy} & Random           & 127,130 (2.5\%)                       & 118,080 (2.3\%)                       \\
                       & GRIT            & 817,991 (16.4\%)                     & 8111,145 (16.2\%)                     \\
\hline
\multirow{2}{*}{4M-Clean} & Random            & 153,006 (3.1\%)                      & 148,729 (3.0\%)                      \\
                       & GRIT            & 1,114,851 (23.2\%)                     & 1,096,485 (22.2\%)      
    \\
\hline
\end{tabular}%
}
\vspace{-.1in}
\label{table:FN_statistics}
\end{table}

\begin{figure}[h!]
   \vspace{-.1in}
   \centering
   \includegraphics[width=\linewidth]{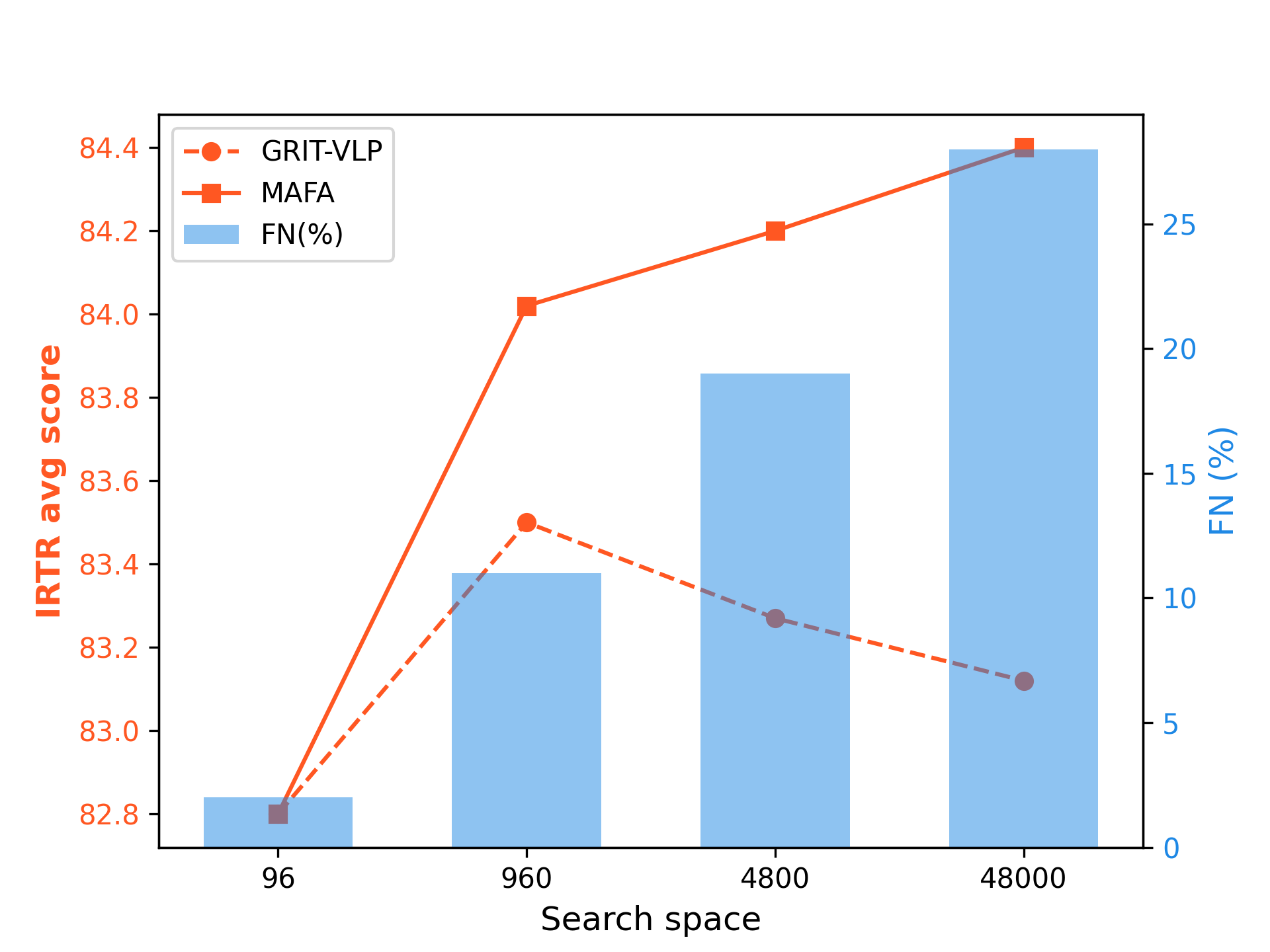}
    \vspace{-.25in}
    \caption{Comparison of IRTR average scores and false negatives (\%) in the 4M-Noisy dataset for GRIT-VLP and MAFA across different search spaces ($M$) when applying GRIT sampling. Here, IRTR average score is defined as the average image-text retrieval accuracy across (TR/R@1, TR/R@5, TR/R@10, IR/R@1, IR/R@5, IR/R@10) on COCO 5k test set.
    For all models, we set the batch size $B$ as $96$. Thus, when $M=96$, GRIT sampling becomes equivalent to random sampling. 
    }
   \label{fig:motivation_plot}
   \vspace{-.2in}
\end{figure}

From the table, we observe that GRIT sampling exhibits significantly more false negatives than random sampling, as mentioned in the Introduction. The reason is that GRIT sampling generates challenging in-batch hard negatives, which are beneficial for learning fine-grained representations, but they also often end up being false negatives. Moreover, we observe this trend exacerbates in the 4M-Clean dataset. 

In Figure \ref{fig:motivation_plot}, we examine the impact of an increasing number of false negatives on the downstream performance of GRIT-VLP. Specifically, we measured the average IRTR score of GRIT-VLP across different search space (queue) sizes $M$, while keeping the batch size $B$ constant. 
We first clearly observe that the number of false negatives rises as $M$ increases. This is expected since expanding the search space for GRIT sampling leads to more similar examples being grouped together in a mini-batch, thereby generating more false negatives.
In terms of the downstream performance of GRIT-VLP, we notice a decline when the value of $M$ exceeds a certain threshold ($M=960$). We attribute this decline to the introduction of ``noise'' caused by the increasing presence of false negatives, which subsequently hampers the effectiveness of hard negative sampling in GRIT-VLP. Based on this analysis, we anticipate that effectively addressing the issue of false negatives while leveraging the potential of hard negative samples will be crucial for enhancing VLP models even further. 
In S.M, we further explore the impact of varying batch sizes on the occurrence of false negatives under the random sampling scenario, which highlights the significance of handling false negatives even in the typical VLP setting (large batch size under random sampling).

In response to this challenge, we propose MAFA, which effectively addresses the issue of false negatives and improves the downstream performance significantly compared to GRIT-VLP. The preview of the performance of MAFA is also shown in Figure \ref{fig:motivation_plot} --- it is evident that the IRTR score of MAFA continues to improve despite an increase in the number of false negatives.

\begin{figure}[t]
   \centering
   \includegraphics[width=\linewidth]{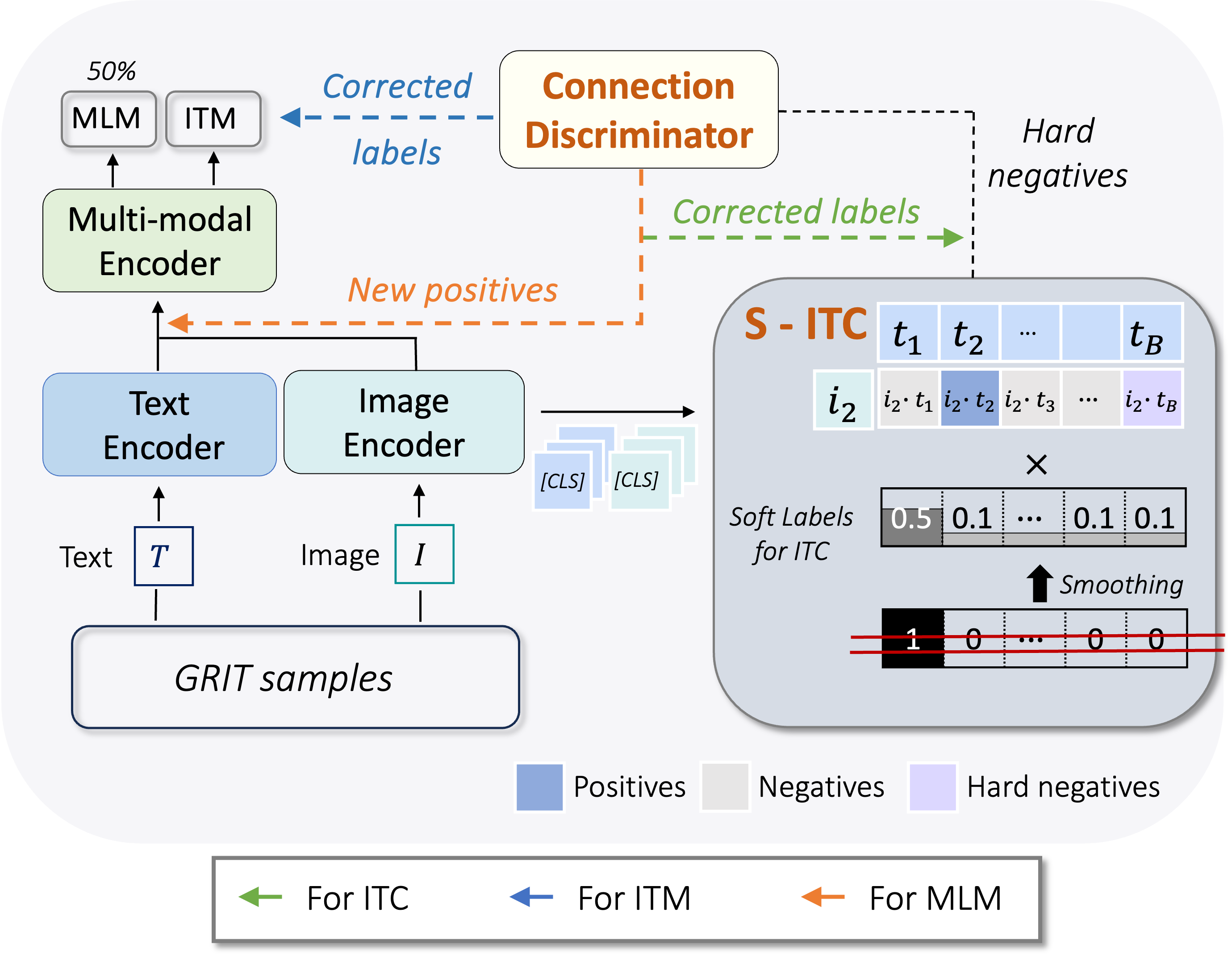}
   \vspace{-.2in}
    \caption{Overall framework of MAFA.}
   \label{fig:overall_MAFA}
   \vspace{-.1in}
\end{figure}
\vspace{-.1in}
\begin{figure*}[t!]
   \centering
   \includegraphics[width=\linewidth]{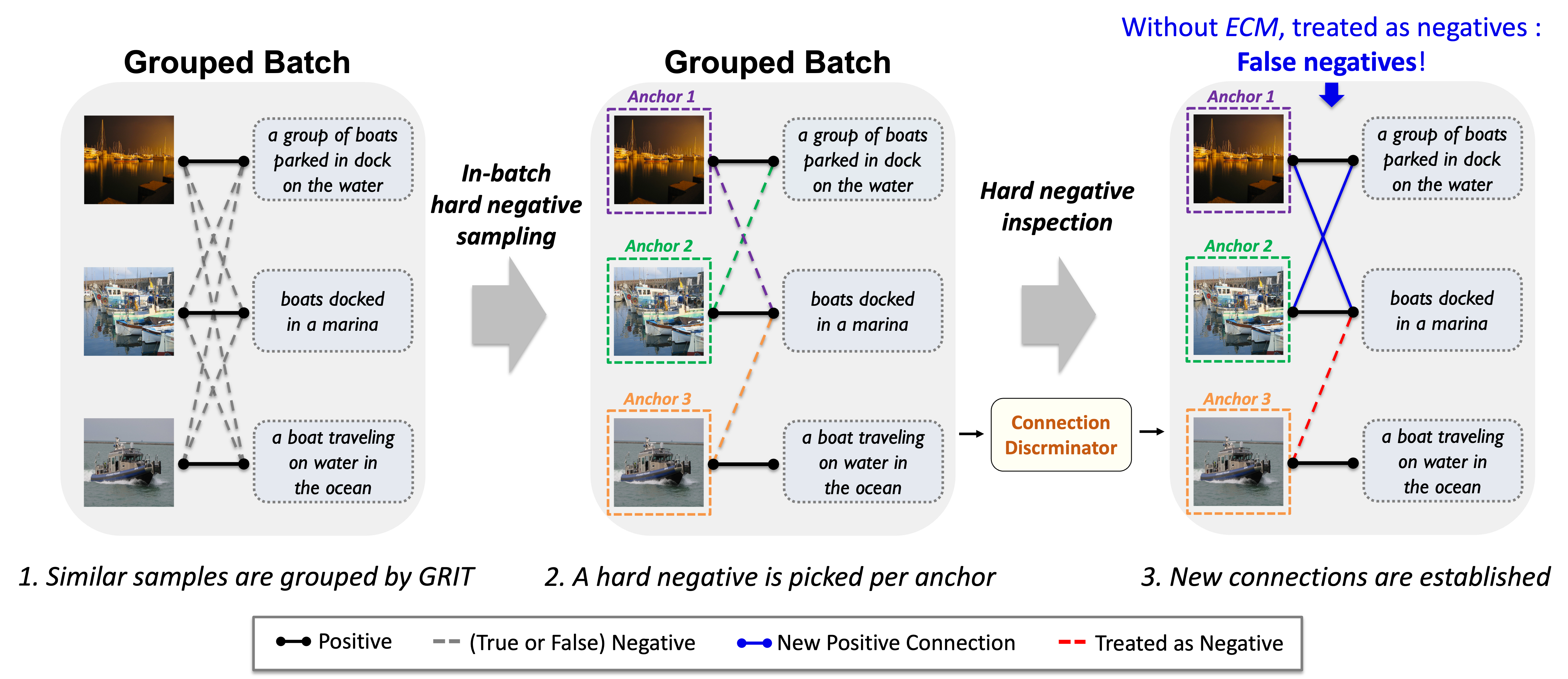}
   \vspace{-.2in}
    \caption{Efficient Connection Mining (ECM).}
    \vspace{-.2in}
   \label{fig:overall_connection}
\end{figure*}
\section{Main Method: MAFA}
\label{sec:method}
Our MAFA consists of two integral components:
we first present the intuition and details of the Efficient Connection Mining (ECM), then delineate the Smoothed ITC (S-ITC). 

\subsection{Efficient Connection Mining (ECM)}
As outlined in the Introduction, the issue of false negatives originates from \textit{missing} positive connections within a paired dataset, a challenge that is computationally infeasible to address naively.
To tackle this, ECM process is strategically designed to exclusively examine hard negatives with significantly higher likelihoods of being false negatives. 
Namely, as described in Figure \ref{fig:overall_connection} and Algorithm 1 in S.M, the training model selects the hardest negatives for each anchor based on ITC similarity in the GRIT-sampled mini-batches.
Once hard negatives are selected, a separate pre-trained ITM model, \textit{Connection Discriminator (Con-D)}, is employed to determine whether these hard negatives are true (hard) negatives or false negatives. 
If \textit{Con-D} assigns a probability to a candidate image-text pair of being positive higher than a threshold $\tau$ (which was set to $0.8$), that candidate is adopted as a new positive pair in the pre-training losses. 

We note that ECM process can be seamlessly integrated with the training process of BLIP-family models (ALBEF \cite{(ALBEF)li2021align}, BLIP \cite{(BLIP)li2022blip}, BLIP-2 \cite{(BLIP-2)li2023blip}) as well by adopting GRIT sampling and \textit{Con-D}, given that ITC and ITM are used as their training objectives.
Moreover, due to the inherent randomness of mini-batches, \textit{Con-D} encounters a variety of hard negatives in each batch and epoch, which enables ECM to create diverse positive connections during training.

Now, we will elaborate on the details of the three pre-training losses (ITC, ITM, and MLM) utilized in our model, and then explain how false negatives identified by ECM are integrated into these losses. 
Briefly, for ITC and ITM, the labels for identified false negatives are revised from negatives to positives. Moreover, these new positives are additionally used as inputs for MLM. 

\noindent\textbf{[ITC with ECM] } In ITC, to measure the similarity between images and texts, the [CLS] tokens from the uni-modal encoders are utilized, as illustrated in Figure \ref{fig:overall_MAFA}. We denote the cosine similarity between image $i$ and text $t$ as $s(i,t)=g_\text{I}(i^{cls})^Tg_\text{T}(t^{cls})$, where $g_\text{I}(\cdot)$ and $g_\text{T}(\cdot)$ are linear projections for [CLS] tokens of image and text embeddings, respectively. The objective of ITC is to maximize the similarity of positive pairs while minimizing that of negative pairs; hence, the loss becomes\vspace{-.1in}
\begin{multline}
\mathcal{L}_{\text{ITC}} = \frac{1}{2}\mathbb{E}_{(i,t)\sim D} \Big[\texttt{CE}\Big(\bm y^{\text{I2T}}(i), \bm p^{\text{I2T}}(i)\Big)\\
+\texttt{CE}\Big(\bm y^{\text{T2I}}(t), \bm p^{\text{T2I}}(t)\Big)\Big],
\label{eq:itc_loss}
\end{multline}
in which $\bm y^{\text{I2T}}(i)$ and $\bm y^{\text{T2I}}(t)$ stand for the one-hot vectors for the correct sample pairs for image $i$ and text $t$, respectively. \texttt{CE}$(\cdot)$ denotes the cross-entropy loss.  The softmax-normalized image-to-text and text-to-image similarities between image $i$ and text $t$, $\bm p_t^{\text{I2T}}(i)$ and $\bm p_i^{\text{T2I}}(t)$, are defined as
\begin{equation}
\bm p_t^{\text{I2T}}(i)=\frac{e^{s(i,t)/\tau}}{\sum_{k=1}^N e^{s(i,t_k)/\tau}}, \bm p_i^{\text{T2I}}(t)=\frac{e^{s(i,t)/\tau}}{\sum_{k=1}^N e^{s(i_k,t)/\tau}}
\label{eq:itc_prob},
\end{equation}
 in which $\tau$ is the temperature and $N$ is the number of considered texts and images. 

To incorporate missing positive connections identified by \textit{Con-D}, the one-hot label $\bm{y}$ is adjusted to $\Tilde{\bm{y}}^{\text{ITC}}$. 
For example, for an anchor image $i$, a single text $t_k$ is picked by in-batch hard negative sampling. Then, if $t_k$ is recognized as a new positive by \textit{Con-D}, the $k$-th element of the one-hot vector $\bm{y}^\text{I2T}(i)$ changes from $0$ to $0.5$. Simultaneously, the original label value of $1$ in $\bm{y}^\text{I2T}(i)$ becomes $0.5$, ensuring that the sum remains $1$. If a positive connection is not newly established, the label remains unchanged. 
The above process is applied identically for text $t$ and image $i_k$. Now, we denote the new ITC loss equipped with $\Tilde{\bm{y}}^{\text{ITC}}$ as $\mathcal{L}_\text{ITC}^\textit{ECM}$.

\noindent\textbf{[ITM with ECM] }  ITM task aims to predict whether the provided pair is matched or not. Similar to ITC, the labels for ITM are revised based on the missing positive connections identified by \textit{Con-D}. 
We employ a re-sampling strategy for the \textit{ambiguous} samples, which are uncertain whether they are false negatives or not (those with a probability of being positive between $0.5$ and $0.8$). 
These ambiguous samples are discarded, and the second hardest text (or image) is re-sampled for the anchor image (or text) to obtain a more certain negative for ITM. Thus, the form of the ITM loss is as follows:
\begin{equation}
\mathcal{L}_{\text{ITM}}^\textit{ECM}=\mathbb{E}_{(i,t)\sim D} \Big[ \texttt{CE}\Big( \Tilde{\bm{y}}^\text{ITM}, \bm{p}^\text{ITM}(i,t)\Big)\Big],
\label{eq:itm_loss}
\end{equation}
in which $\Tilde{\bm{y}}^\text{ITM}$ represents the corrected one-hot label from \textit{Con-D}. 

\noindent\textbf{[MLM with ECM] } For MLM, the model is asked to predict masked tokens in the caption using unmasked text tokens and visual information. In addition to the original positive pairs in the dataset, new positive pairs detected by \textit{Con-D} are additionally used:
\begin{equation}
\mathcal{L}_{\text{MLM}}^\textit{ECM}=\mathbb{E}_{(i,t)\sim D\cup D_\text{ECM}}\Big[ \texttt{CE}\Big( \bm{y}^\text{MLM}, \bm{p}^\text{MLM}(i,t^\text{mask})\Big)\Big],
\label{eq:mlm_loss}
\end{equation}
in which $D_\text{ECM}$ denotes the set of newly constructed pairs, $\bm{y}^\text{MLM}$ represents the one-hot label for the masked token, and $\bm{p}^\text{MLM}(i,t^\text{mask})$ indicates the model's prediction for the masked token.


\noindent\textit{Remark: }  
\textit{Con-D} is pre-trained with the following pre-training objectives: S-ITC, MLM, and ITM with GRIT sampling. Then, it is fine-tuned on the Karpathy training split of MS-COCO \cite{(IRTR)lin2014microsoft}, and the output of ITM head of \textit{Con-D} serves as the probability for a candidate image-text pair to be a positive pair. The additional computation overhead of ECM during training depends only on the number of samples provided to \textit{Con-D} and the number of new positives given to the multi-modal encoder of the training model for MLM. 
Since only labels are corrected for ITC and ITM, it does not require additional forward passes for the model being trained.
Thus, despite the inclusion of additional forwarding passes for ECM, the extra overhead introduced by ECM is relatively low compared to the momentum distillation technique employed in ALBEF and BLIP.
Detailed information regarding the computational cost is described in S.M.


\subsection{Smoothed ITC (S-ITC) }
\label{sec:method_UITC}
To overcome the challenge of false negatives in ITC under GRIT sampling, we additionally introduce a computation-free approach named S-ITC, which employs label smoothing to contrastive loss, which has not been extensively explored in VLP.
Specifically, we take the following loss form:
 \begin{multline}
     \mathcal{L}_{\text{S-ITC}} = \frac{1}{2}\mathbb{E}_{(i,t)\sim D }\Big[\texttt{CE}\Big((1-\alpha)\bm y^{\text{I2T}}(i) + \frac{\alpha}{N}\bm{1}, \bm p^{\text{I2T}}(i)\Big) \\
+\texttt{CE}\Big((1-\alpha)\bm y^{\text{T2I}}(t) + \frac{\alpha}{N}\bm{1}, \bm p^{\text{T2I}}(t)\Big)\Big], 
\label{eq:s_itc_loss}
 \end{multline}
 in which $\alpha$ represents a mixing parameter, and $\bm{1}$ denotes all-one vector. 

 We emphasize that label smoothing has not been widely adopted in typical VLP settings since it has not been very effective. 
 As we show in Table \ref{table:ITC-ablation} (Section \ref{sec:experiments}), performance is significantly degraded when S-ITC is applied under the random sampling scenario.
 This decline is largely due to the detrimental effect of providing soft labels for the examples in the randomly sampled batch where true negatives are prevalent.
 In contrast, under GRIT sampling where each mini-batch is predominantly composed of samples that are likely to be false negatives, we observe that S-ITC, which ensures relatively high soft labels for all negatives, becomes highly effective.

There also have been other attempts to address the issue of false negatives in ITC, such as momentum distillation \cite{(ALBEF)li2021align} and consistency loss \cite{(GRIT-VLP)byun2022grit}. 
Here, we explain only I2T-related terms for simplicity; T2I-related terms are similarly computed. Momentum distillation replaces $\bm y_t^{\text{I2T}}(i)$ by $\bm y_t^{\text{MD}}(i)=(1-\alpha)\bm y_t^{\text{I2T}}(i)+\alpha sg[\Tilde{\bm p}_t^{\text{I2T}}(i)]$, where $sg[\cdot]$ is the stop gradient operator, and $\Tilde{\bm p}$ denotes the probability obtained from the momentum encoder. Here, $N$ is equal to batch-size $B+Q$ since the model is accompanied by a queue of size $Q$ that stores embeddings to provide additional negatives. However, this approach suffers from inefficiency due to the additional forwarding of the momentum model, and it results in increasing the model size. In consistency loss, $\bm y_t^{\text{I2T}}(i)$ is substituted with $\bm y_t^{\text{CS}}(i)=(1-\alpha)\bm y_t^{\text{I2T}}(i)+\alpha sg[\bm p_i^{\text{T2I}}(t)]$, where $\bm p_i^{\text{T2I}}(t)$ is computed by the model itself. Here, $N$ is the same as $B$ since it does not involve a queue.
\begin{table}[]
\caption{Values of soft labels assigned to samples in ITC for different methods. The batch size $B$ is set to 96, and the queue size $Q$ is set to 48000. The soft labels were computed in the last epoch of the training.}
\vspace{-.1in}
\centering
\resizebox{\columnwidth}{!}{%
\begin{tabular}{c||c|c|c}
\hline
\multirow{2}{*}{Method} & \multicolumn{3}{c}{Sum of soft labels}                                                  \\ \cline{2-4} 
                        & Top 1 $\sim$ 5 & Top $6\sim B$ & Top $B+1\sim B+Q$ \\ \hline
S-ITC                   & 0.5260        & 0.4740        & $\cdot$           \\ \hline
Consistency Loss            & 0.9822        & 0.0178        & $\cdot$           \\ \hline
Momentum Distillation                      & 0.6746        & 0.0009        & 0.3245       \\ \hline
\end{tabular}%
}
\vspace{-.1in}
\label{table:sum_of_soft_labels}
\end{table}

However, as shown in Table \ref{table:ITC-ablation} (in Section \ref{sec:experiments}), the effectiveness of momentum distillation and consistency loss is limited.
To explore the reason behind this, we examine the soft labels from the above methods in the GRIT sampling scenario as reported in Table \ref{table:sum_of_soft_labels}, aiming to uncover the distribution shapes of the soft labels.
The values in the table are obtained through the following process: the soft labels are sorted in descending order, and then averaged across all samples. Further details on the computation process are described in S.M.
We observe that momentum distillation continues to assign almost zero labels to negative samples, which are likely to be false negatives under GRIT sampling.
This result may stem from the large number of negatives in the queue, which prevents each negative sample from receiving non-negligible labels. 
On the other hand, consistency loss assigns comparatively higher soft labels (0.0178) than momentum distillation (0.0009) but overly concentrates on a few pairs, resulting in negligible labels for most negatives.
In S.M, we provide an analysis that this phenomenon cannot be resolved by merely tuning $\alpha$.

Given that both consistency loss and the momentum distillation fail to achieve the intended objective of assigning non-negligible soft labels to the majority of negatives, we argue that S-ITC, which explicitly assigns higher soft labels to all negatives, can be a simple but effective solution.
In S.M, we include an analysis of its robustness to $\alpha$. 

Consequently, as illustrated in Figure \ref{fig:overall_MAFA} and Algorithm 1 in S.M, we adopt the following pre-training objective:
\begin{equation}
\mathcal{L} = \mathcal{L}_{\text{S-ITC}}^\textit{ECM}+ \mathcal{L}_{\text{MLM}}^\textit{ECM} + \mathcal{L}_{\text{ITM}}^\textit{ECM}  ,
\label{eq:overall_loss}
\end{equation}
where $\mathcal{L}_{\text{S-ITC}}^\textit{ECM}$ represents the integrated ITC loss of $\mathcal{L}_{\text{ITC}}^\textit{ECM}$ and $\mathcal{L}_{\text{S-ITC}}$, which adopts the target labels as $(1-\alpha)\Tilde{\bm{y}}^\text{ITC} + \frac{\alpha}{N}\bm{1}$.

%% file: 04_experiments.tex
\begin{table*}[]
\caption{Comparison with various methods on downstream vision-language tasks. \textbf{Bold} denotes the best result among models trained with 4M dataset. * refers to the reproduced models by the authors. Methods without explicit designation are trained on 4M-Noisy dataset.}
\vspace{-.1in}
\resizebox{\linewidth}{!}{%
\begin{tabular}{ccc|ccccccccccc}
\hline
\multirow{2}{*}{Method} & \multicolumn{1}{l}{} & \multicolumn{1}{l|}{\multirow{2}{*}{\begin{tabular}[c]{@{}l@{}}Pre-train\\ \# Images\end{tabular}}} & \multicolumn{2}{c}{COCO R@1} &  & \multicolumn{2}{c}{Flickr R@1} &  & \multicolumn{2}{c}{NLVR2} &  & \multicolumn{2}{c}{VQA} \\
                        & \multicolumn{1}{l}{} & \multicolumn{1}{l|}{}                                                                               & TR            & IR           &  & TR             & IR            &  & dev         & test-P      &  & test-dev   & test-std   \\ \hline
UNITER \cite{(UNITER)chen2020uniter}                 &                      & 4M                                                                                                  & 65.7          & 52.9         &  & 87.3           & 75.6          &  & 77.18       & 77.85       &  & 72.70      & 72.91      \\
VILLA \cite{(VILLA)gan2020large}                  &                      & 4M                                                                                                  & -             & -            &  & 87.9           & 76.3          &  & 78.39       & 79.30       &  & 73.59      & 73.67      \\
OSCAR \cite{(OSCAR)li2020oscar}                  &                      & 4M                                                                                                  & 70.0          & 54.0         &  & -              & -             &  & 78.07       & 78.36       &  & 73.16      & 73.44      \\
ALBEF \cite{(ALBEF)li2021align}                  &                      & 4M                                                                                                  & 73.1          & 56.8         &  & 94.3           & 82.8          &  & 80.24       & 80.50       &  & 74.54      & 74.70      \\
TCL \cite{(TCL)yang2022vision}                    &                      & 4M                                                                                                  & 75.6          & 59.0           &  & 94.9           & 84.0            &  & 80.54       & 81.33       &  & 74.90      & 74.92      \\
BLIP* (4M-Clean) \cite{(BLIP)li2022blip}                  &                      & 4M                                                                                                  & 75.5       &  58.9     &  & 94.3           & 82.6         &  &  79.70           & 80.87            &  &  75.50          &  75.76         \\
GRIT-VLP* \cite{(GRIT-VLP)byun2022grit}               &                      & 4M                                                                                                  & 76.6          & 59.6         &  & 95.5           & 82.9          &  & 81.40       & 81.23       &  & 75.26      & 75.32      \\ \hline
\textbf{MAFA}                    &                      & 4M                                                                                                  & 78.0          & 61.2         &  & 96.1           & \textbf{84.9}          &  & 82.52       & 82.08       &  & 75.55      & 75.75      \\
\textbf{MAFA} (4M-Clean)         &                      & 4M                                                                                                  & \textbf{79.4}          & \textbf{61.6}         &  & \textbf{96.2}           & 84.6          &  & \textbf{82.66}       & \textbf{82.16}       &  & \textbf{75.91}      & \textbf{75.93}      \\ \hline
ALBEF                   &                      & 14M                                                                                                 & 77.6          & 60.7         &  & 95.9           & 85.6          &  & 82.55       & 83.14       &  & 75.85      & 76.04      \\
BLIP                    &                      & 14M                                                                                                 & 80.6          & 63.1         &  & 96.6           & 87.2          &  & 82.67       & 82.30       &  & 77.54      & 77.62     
\end{tabular}
}
\vspace{-.1in}
\label{tab:SOTA}
\end{table*}
\begin{table*}[]
\caption{Ablation study on the proposed method. \textbf{Bold} denotes the best result among models trained with 4M-Noisy, 4M-Clean dataset, respectively.}
\vspace{-.2in}
\center{
\resizebox{0.9\linewidth}{!}{%
\begin{tabular}{c|cc||llcllcllcll}
\hline
\multirow{2}{*}{\begin{tabular}[c]{@{}c@{}}Pre-train\\ dataset\end{tabular}} & \multicolumn{2}{c||}{MAFA}               & \multicolumn{2}{c}{COCO R@1}                    &  & \multicolumn{2}{c}{Flickr R@1}                      &  & \multicolumn{2}{c}{NLVR2}                            &  & \multicolumn{2}{c}{VQA}                                     \\ \cline{2-3}
                                                                             & \multicolumn{1}{c|}{S-ITC} & ECM & \multicolumn{1}{c}{TR} & \multicolumn{1}{c}{IR} &  & \multicolumn{1}{c}{TR}   & \multicolumn{1}{c}{IR}   &  & \multicolumn{1}{c}{dev} & \multicolumn{1}{c}{test-P} &  & \multicolumn{1}{c}{test-dev} & \multicolumn{1}{c}{test-std} \\ \hline
\multirow{4}{*}{4M-Noisy}                                                       & \multicolumn{1}{c|}{\xmark }         & \xmark      & 76.6                  & 59.6                  &  & 95.5                     & 82.9                     &  & 81.40                   & 81.23                      &  & 75.26                        & 75.32                        \\
                                                                             & \multicolumn{1}{c|}{\xmark}         & \cmark      & 77.4                  & 60.2                  &  & 95.5                     & 83.3                     &  & 82.03                   & 81.76                      &  & 75.39                        & 75.52                        \\
                                                                             & \multicolumn{1}{c|}{\cmark}         & \xmark      & 77.5                  & 60.5                  &  & 96.1                     & 84.2                     &  & 81.74                   & 81.33                      &  & 75.42                        & 75.51                        \\
                                                                             & \multicolumn{1}{c|}{\cmark}         & \cmark      & \textbf{78.0}                  & \textbf{61.2}                  &  & \textbf{96.1}                     & \textbf{84.9}                     &  & \textbf{82.52}                   & \textbf{82.08}                      &  & \textbf{75.55}                        & \textbf{75.75}                        \\ \hline
\multirow{2}{*}{4M-Clean}                                                       & \multicolumn{1}{c|}{\xmark}         & \xmark      & 77.7                  & 60.7                  &  & \multicolumn{1}{c}{95.2}     & \multicolumn{1}{c}{84.2}     &  & 81.44                   & 81.39                      &  & 75.50                        & 75.57                        \\
                                                                             & \multicolumn{1}{c|}{\cmark}         & \cmark      & \textbf{79.4}                  & \textbf{61.6}                  &  & \multicolumn{1}{c}{\textbf{96.2}} & \multicolumn{1}{c}{\textbf{84.6}} &  & \textbf{82.66}                   & \textbf{82.16}                      &  & \textbf{75.91}                        & \textbf{75.93}                        \\ \hline
\end{tabular}
}
\vspace{-.1in}
\label{tab:main-analysis}
}
\end{table*}

\section{Experimental Results} \label{sec:experiments}
\subsection{Data and experimental settings}
During our training process, we utilize four datasets (MS-COCO \cite{(IRTR)lin2014microsoft}, Visual Genome \cite{(VG)krishna2017visual}, Conceptual Captions \cite{(CC)sharma2018conceptual}, and SBU Captions \cite{(SBU)ordonez2011im2text}) with a total of 4M unique images (5M image-text pairs), as proposed by ALBEF \cite{(ALBEF)li2021align} and UNITER \cite{(UNITER)chen2020uniter}. 
We refer to this collective dataset as the ``4M-Noisy'' dataset due to a significant number of captions that offer either incomplete or incorrect descriptions, which can be seen as false positives.
To analyze the impact of our approach in handling false negatives relative to the effect of removing false positives, we construct an additional same-sized training set named  ``4M-Clean'' which is composed of clean image-text pairs, refined by the BLIP captioner \cite{(BLIP)li2022blip}. 
Note that all the models are pre-trained with the ``4M-Noisy'' unless specifically stated as ``4M-Clean'' in our results table below.
Further details on constructing the 4M-Clean dataset are in S.M.

Following ALBEF, we adopt our image encoder as a 12-layer Vision Transformer \cite{(VIT)dosovitskiy2020image}  with 86 million parameters, pre-trained on ImageNet-1k \cite{(imagenet1K)touvron2021training}. 
Both the text and multi-modal encoders utilize a 6-layer Transformer \cite{(attention)vaswani2017attention}, initializing the former with the first 6 layers and the latter with the last 6 layers of BERT-base model (123.7M parameters) \cite{(BERT)devlin2018bert}.
We use the same data augmentation method used in ALBEF and train our model for 20 epochs using 4 NVIDIA A100 GPUs, but excluding the momentum encoder in ALBEF.
For \textit{Con-D}, we use the exact same model architecture as the training model. 
Unless otherwise noted, we set $B = 96$ and $M = 4800$ for GRIT sampling, and for all other hyper-parameter settings, we follow GRIT-VLP \cite{(GRIT-VLP)byun2022grit}.
More details on the dataset, training, and hyperparameters are in S.M.

\subsection{Downstream vision and language tasks}
After completing the pre-training phase, we proceed to fine-tune our model on three downstream vision and language tasks: image-text retrieval (IRTR) \cite{(IRTR)lin2014microsoft}, visual question answering (VQA) \cite{(VQAv2)goyal2017making}, and natural language for visual reasoning (NLVR2) \cite{(NLVR)suhr2018corpus}. For IRTR, we utilize the MS-COCO \cite{(IRTR)lin2014microsoft} and Flickr30K (F30K) \cite{(Flickr)plummer2015flickr30k} datasets, with F30K being re-splitted according to \cite{(Flickr_split)karpathy2015deep}. 
Following BLIP \cite{(BLIP)li2022blip}, we exclude the SNLI-VE dataset \cite{(SNLI-VE)xie2019visual} due to reported noise in the data.
Our fine-tuning and evaluation process mostly follows that of GRIT-VLP. 
More details of downstream tasks are in S.M.

\subsection{Comparison with baselines}
In  Table \ref{tab:SOTA},  we observe that our approach consistently outperforms other baselines in multiple downstream tasks (IRTR, VQA, NLVR2).
Notably, MAFA even surpasses ALBEF (14M) and competes with BLIP (14M) on certain metrics, despite being trained on a significantly smaller dataset.
Specifically, MAFA achieves significant improvements over GRIT-VLP, with a substantial margin of $+1.4\%$ IR/R@1, $+1.6\%$ TR/R@1 on MS-COCO and $+1.1\%$ on NLVR2 dev. These results clearly show the significance of addressing false negatives when leveraging hard negative mining.
Additionally, we believe that the comparison between BLIP (4M-Clean) and our MAFA shows that the effectiveness of managing false negatives may surpass the impact of mitigating false positives.
Furthermore, the enhanced performance of MAFA (4M-Clean) over MAFA shows the synergistic effect of addressing both false positives and negatives.

\subsection{Ablation studies}
Table \ref{tab:main-analysis} presents the effectiveness of two proposed components: efficient connection mining (ECM) and smoothed ITC (S-ITC). 
Here, all model variants adopt GRIT sampling, with row 1 representing the original GRIT-VLP model.
The results clearly demonstrate that applying either the S-ITC (row 3) or the ECM (row 2) individually leads to performance improvements compared to a model that does not consider false negatives (row 1). 
By combining both S-ITC and ECM in our final model (row 4), we observe significant performance enhancements on the 4M-Noisy dataset. %
This tendency is validated again in the 4M-Clean dataset, confirming the consistent effectiveness of the proposed components (row 6).
Moreover, by comparing the performance gap between MAFA trained on the noisy dataset (row 4) and GRIT-VLP trained on the clean dataset (row 5), we reaffirm that addressing false negatives outweighs the impact of handling false positives. Beyond the 4M dataset, we present additional results across a broader range of data scales (1M, 2M, and 14M) in S.M, demonstrating the robustness of MAFA with respect to data scale variations.




\begin{table}[]
\caption{Analysis of the effect of MAFA with GRIT sampling. ``ECM-E''  denotes eliminating false negatives rather than using them as positives. }
\vspace{-.1in}
\resizebox{\linewidth}{!}{
\begin{tabular}{cc||cc|cc|cc}
\hline
\multicolumn{2}{c||}{Method}       & \multicolumn{2}{c|}{COCO R@1} & \multicolumn{2}{c|}{NLVR2} & \multicolumn{2}{c}{VQA} \\ \cline{1-2}
\multicolumn{1}{c|}{GRIT} & MAFA & IR            & TR            & dev         & test-P       & test-dev   & test-std   \\ \hline
\multicolumn{1}{c|}{ \xmark }    &  \xmark      & 74.4          & 57.6          & 79.75       & 79.94        & 74.49      & 74.67      \\
\multicolumn{1}{c|}{ \xmark }    & \cmark     & 74.3          & 57.8          & 81.20       & 81.03        & 74.61           & 74.78       \\ \hline
\multicolumn{1}{c|}{ \cmark}    &  \xmark      & 76.6          & 59.6          & 81.40       & 81.23        & 75.26      & 75.32      \\
\multicolumn{1}{c|}{ \cmark}    & \cmark (ECM-E)     & 77.1          & 61.1          & 82.33           & 81.95             & 75.50      & 75.54      \\
\multicolumn{1}{c|}{ \cmark}    & \cmark     & \textbf{78.0}          & \textbf{61.2}          & \textbf{82.52}       & \textbf{82.08}        & \textbf{75.55}      & \textbf{75.75}      \\ \hline
\end{tabular}
}
\vspace{-.1in}
\label{table:with_GRIT}
\vspace{-.1in}
\end{table}

Table \ref{table:with_GRIT} provides an additional comparative analysis on the effectiveness of MAFA, based on whether GRIT sampling and ECM are either applied or not.
Here, row 1 denotes the ALBEF model without momentum distillation.
Since S-ITC is ineffective under random sampling (as we show in Table \ref{table:ITC-ablation} below), S-ITC is excluded when GRIT sampling is not utilized (row2).
We observe that MAFA enhances the performance for both random and GRIT sampling. 
However, the effect of MAFA is much more vivid for GRIT sampling (row 4, 5), underscoring the critical role of managing false negatives in hard negative sampling.
Moreover, our experiments reveal that converting false negatives into additional positives (row 5) is considerably more beneficial than merely removing them (row 4), which highlights the effect of leveraging new positive connections constructed by the model within the dataset itself.

Furthermore, in Table \ref{table:ITC-ablation}, we provide a comparative analysis on S-ITC, which supports our discussion in Section \ref{sec:method_UITC}; S-ITC brings out a unique synergy only when combined with GRIT-sampling.
Namely, in random sampling, we observe that S-ITC rather detrimentally affects performance (row 2).
Conversely, under GRIT sampling, we verify that assigning relatively high nonzero labels to most negatives enhances performance. 
Namely, consistency loss (row 4), which assigns relatively higher soft labels to samples in a batch, outperforms momentum distillation (row 5). 
S-ITC significantly outperforms the other two variants, which highlights the importance of assigning substantial labels to the majority of negatives, rather than just a few.

\subsection{Compatibility of MAFA with BLIP-2 \cite{(BLIP-2)li2023blip}}
In Tables \ref{table:with_BLIP-2} and \ref{tab:BLIP-2-zero}, we demonstrate that our MAFA can be successfully integrated with the recent BLIP-2 \cite{(BLIP-2)li2023blip}, which is quite a successful vision-language pre-training framework.
As described in S.M., the \textit{stage-1} of BLIP-2, which adopts ITC, ITM, and (auto-regressive) LM losses as objectives, closely resembles the pre-training procedures of both BLIP and ALBEF. Thus, MAFA can be effortlessly incorporated into \textit{stage-1} of BLIP-2, following the identical way described in Section \ref{sec:method}.

In Table \ref{table:with_BLIP-2}, we observe the performance of BLIP-2+GRIT is significantly degraded (row 2), which indicates that solely applying GRIT sampling leads to a failure of learning. 
We believe this performance degradation primarily stems from more frequent occurrences of false negatives in BLIP-2. 
In BLIP-2, due to the significantly enhanced capacity of the model, GRIT sampling, which mines hard negatives based on contrastive similarities calculated from the training model, includes a substantially higher number of false negatives in each batch. 
The integration of MAFA with BLIP-2 leads to enhanced performance, highlighting the importance of managing false negatives to increase the stability of the training process.

\begin{table}[t!]
\caption{ Comparison of soft-labeling methods for ITC. }
\vspace{-.1in}
\resizebox{\columnwidth}{!}{%
\begin{tabular}{cc||cc|cc}
\hline
\multicolumn{2}{c||}{Method}                                     & \multicolumn{2}{c|}{COCO R@1}   & \multicolumn{2}{c}{Flickr R@1} \\ \cline{1-2}
\multicolumn{1}{c|}{GRIT}               & Soft labeling         & TR             & IR             & TR            & IR             \\ \hline
\multicolumn{1}{c|}{\multirow{3}{*}{\xmark}} & \xmark                     & \textbf{74.4}  & \textbf{57.6} & \textbf{93.5} & 81.7          \\
\multicolumn{1}{c|}{}                   & Momentum Distillation & 74.2           & 57.4          & \textbf{93.5} & \textbf{81.9} \\
\multicolumn{1}{c|}{}                   & S-ITC                 & 73.5          & 56.1          & 92.9          & 79.9          \\ \hline
\multicolumn{1}{c|}{\multirow{3}{*}{\cmark}} & Consistency Loss      & 76.6          & 59.6          & 95.5          & 82.9           \\
\multicolumn{1}{c|}{}                   & Momentum Distillation & 76.1          & 58.9          & 94.4          & 82.7           \\
\multicolumn{1}{c|}{}                   & S-ITC                 & \textbf{77.5} & \textbf{60.5} & \textbf{96.1} & \textbf{84.2}  \\ \hline
\end{tabular}
}
\vspace{-.1in}
\label{table:ITC-ablation}
\end{table}

\begin{table}[t!]
\caption{ Fine-tuned IRTR results with BLIP-2 framework on MS-COCO datasets.}
\vspace{-.1in}
\resizebox{\linewidth}{!}{
\begin{tabular}{c|ccccccc}
\hline
\multirow{2}{*}{Model}  & \multicolumn{3}{c}{TR} & \multicolumn{3}{c}{IR} \\
                    & R@1    & R@5   & R@10  & R@1    & R@5   & R@10  \\ \hline
BLIP-2               & 82.6       & 96.3      & 98.2      & 66.1       & 86.8      & \textbf{92.0}      \\
BLIP-2 + GRIT        & 65.9       & 89.1      & 95.0      & 52.5       & 79.4      & 87.3      \\
BLIP-2 + MAFA        & \textbf{83.7}       & \textbf{96.6}      & \textbf{98.4}      & \textbf{66.7}       & \textbf{86.8}      & 91.9      \\ \hline
\end{tabular}
}
\label{table:with_BLIP-2}
\vspace{-.1in}
\end{table}
We further explore whether the integration of MAFA in \textit{stage-1} leads to improved generative learning capabilities after additional \textit{stage-2} training where the model is connected to a frozen LLM and pre-trained only with LM loss.
We evaluate the performance of models in various zero-shot visual question answering benchmark datasets including GQA \cite{(GQA)hudson2019gqa}, OKVQA \cite{(OKVQA)marino2019ok}, and VQA \cite{(VQAv2)goyal2017making}. 
Moreover, we assess the zero-shot image captioning ability on the Karpathy test split of MS-COCO \cite{(IRTR)lin2014microsoft}.
Table \ref{tab:BLIP-2-zero} shows that MAFA significantly improves zero-shot performance across various VQA and image captioning tasks.
This result not only underscores the compatibility of MAFA with BLIP-2 but also emphasizes that the exclusive integration of MAFA in \textit{stage-1} is also beneficial in generative learning capability (\textit{stage-2}) as well.
More detailed results, including those from fine-tuned image captioning and an analysis on how extra positive examples from ECM influence the BLIP-2 \textit{stage-2} performance, are provided in S.M.

\begin{table}[t!]
\caption{ Zero-shot visual question answering and image captioning results with BLIP-2 framework. }
\vspace{-.1in}
\resizebox{\columnwidth}{!}{%
\begin{tabular}{c|c|c|c|cc}
\hline
\multirow{2}{*}{Model} & VQAv2 & OK-VQA & GQA      & \multicolumn{2}{c}{\begin{tabular}[c]{@{}c@{}}COCO zero-shot\\ Karpathy test\end{tabular}} \\
                       & val   & test   & test-dev & BLEU@4                                       & CIDEr                                        \\ \hline
BLIP-2                  & 46.6  & 23.8   & 29.1     & 35.6                                        & 118.8                                        \\ \hline
BLIP-2 + MAFA          & \textbf{50.8}  & \textbf{29.0}   & \textbf{31.8}     & \textbf{37.6}                                        & \textbf{125.4}                                        \\
\hline
\end{tabular}
}
\vspace{-.1in}
\label{tab:BLIP-2-zero}
\end{table}

%% file: 10_conclusion.tex
\vspace{-.1in}
 
\section{Concluding Remarks}
We introduce MAFA, a novel approach equipped with two key components (ECM, S-ITC), specifically designed to tackle the prevalent issue of false negatives in VLP. 
Our comprehensive experiments demonstrate that addressing false negatives plays a crucial role in VLP. Moreover, we believe that the concept of converting false negatives into additional positives paves the way for future research that leverages the inherent missing positive connections within a dataset. 
\vspace{-.05in}
\section*{Acknowledgment}
\vspace{-.05in}
This work was supported in part by the National Research Foundation of Korea (NRF) grant [No.2021R1A2C2007884] and by Institute of Information \& communications Technology Planning \& Evaluation (IITP) grants
[No.2021-0-01343, No.2021-0-02068, No.2022-0-00113,
No.2022-0-00959] funded by the Korean government (MSIT). It was also supported by SNU-Naver Hyperscale AI Center.

\newpage



%% file: 12_appendix.tex
\section{Data and Implementation Details} \label{sec:appendix_data}
In this section, we provide information about the software and the dataset used in our study.
We conducted experiments with four NVIDIA A100 GPUS with Python 3.8 and Pytorch with CUDA 11.1.

To construct the 4M-Clean dataset, we download the dataset corpus (CC3M+CC12M+SBU) from the official github of BLIP. The dataset corpus is generated and filtered by a model equipped with VIT-B/16 as its image transformer. 
From this corpus, we selectively utilize image-text pairs from the (CC3M+SBU) to align with the 4M-Noisy dataset. 
Note that the dataset refinement process (generating clean captions, and filtering noisy image-text pairs) is exclusively done with the web-crawled dataset (CC3M+SBU), and (COCO+VG) dataset is incorporated in the 4M-Clean dataset without any modification.

As described in the manuscript (Section \ref{sec:experiments}), we mainly use a synthetically generated and filtered set $(I_w,T_s)$ from this dataset corpus.
To ensure that the size of the 4M-Clean dataset is nearly identical to the 4M-Noisy dataset, we additionally use the small number of $(I_w,T_w)$ among the (CC3M+SBU) dataset due to the synthetically generated and filtered set $(I_w,T_s)$ being slightly smaller than the 4M-Noisy dataset.
The total number of image-text pairs in the 4M-Noisy set and the 4M-Clean set is $4,999,065$ and $4,933,639$, respectively.

\section{Details on counting false negatives in Section \ref{sec:motivation}  (manuscript)}
Here, we first clarify the purpose of using the pre-trained BLIP (129M) model in quantitative analysis in Section \ref{sec:motivation} (manuscript) for better understanding. The quantitative analysis represented in Table \ref{table:FN_statistics} and Figure \ref{fig:motivation_plot} (manuscript) includes the estimated count of false negative pairs during the ITM task of training. 
The main goal is to compare the number of false negatives arising from randomly constructed mini-batches (typical VLP setting) to those from GRIT-sampled mini-batches that group similar pairs in each mini-batch. 
Here, to identify false negative pairs with perfect accuracy, it is essential to employ a human evaluation process, requiring manual examination of each individual negative pair (constructed for performing ITM task) to determine if it is matched or not. 
However, given the infeasibility of manually checking all negative pairs, we have opted to leverage the pre-trained BLIP (129M) model, a strong ITM model, as an alternative to human evaluation. 
While the BLIP (129M) does not always classify with perfect accuracy, it is deemed sufficiently reliable for approximating the tendency of the number of false negatives.

To validate this, we additionally conduct a human evaluation on randomly sampled 200 false negatives in the 4M-Clean dataset during training which are filtered by BLIP (129M) filter. For this, two ML researchers manually check whether each false negative is genuinely a false negative or not. 
In this analysis, among the samples classified as false negatives by BLIP (129M), over 83\% are also determined to be false negatives upon human evaluation. While it's not 100\% accurate, we believe that using BLIP (129M) is reasonable for approximating the number of false negatives during training and, consequently, for comparing the occurrence of false negatives between random sampling and GRIT sampling. To report representative numbers, the values provided in Table \ref{table:FN_statistics} (manuscript) are obtained by multiplying the actual values from the BLIP (129M) model by 0.83, assuming that human correction is statistically significant. The raw values prior to multiplication are presented in rows 1 and 2 of Table \ref{table:FN_statistics_many_discriminator}.

Additionally, we examine the number of false negatives using other discriminators, as detailed in Table \ref{table:FN_statistics_many_discriminator}. Unlike the values in Table \ref{table:FN_statistics} (manuscript), these values are not multiplied by 0.83. ALBEF (14M) denotes the ALBEF model pre-trained on the 14M dataset and fine-tuned on the MS-COCO dataset, while BLIP-2 represents the BLIP-2 model (equipped with ViT-g) pre-trained on 129M and fine-tuned on the MS-COCO dataset. These results also indicate that the tendency of false negative ratios remains consistent across different discriminators.







\begin{table}[h!]
\caption{ The number of false negatives (FNs) counted by various discriminators. Unlike Table \ref{table:FN_statistics} (manuscript), these values represent the raw counts without any adjustments, as the human correction factor of 0.83 has not been applied.}
\vspace{-.1in}
\resizebox{\columnwidth}{!}{%
\begin{tabular}{c|c|c||c|c}
\hline
\textbf{Discriminator} & \textbf{Dataset}      & \textbf{Sampling} & \textbf{FN w.r.t. image} & \textbf{FN w.r.t. text} \\ \hline
\multirow{4}{*}{BLIP (129M)} & \multirow{2}{*}{4M-Noisy} & Random           & 146,127 (2.9\%)                       & 142,265 (2.8\%)                       \\
                      &  &GRIT            & 985,531 (19.7\%)                     & 977,283 (19.5\%)                     \\

\cline{2-5}
 &\multirow{2}{*}{4M-Clean}  & Random           & 184,345 (3.7\%)                       & 179,191 (3.6\%)                       \\
                       & &GRIT            & 1,383,752 (28.0\%)                     & 1,321,066 (26.8\%)                     \\

\hline
 \multirow{2}{*}{ALBEF (14M)}& \multirow{2}{*}{4M-Clean} & Random           & 216,179 (4.4\%)                       & 210,976 (4.3\%)                       \\
                       & &GRIT            & 1,565,023 (31.7\%)                     & 1,477,597 (29.9\%)                     \\

\hline
 \multirow{2}{*}{BLIP-2} & \multirow{2}{*}{4M-Clean} & Random            & 118,118 (2.4\%)                      & 115,281 (2.3\%)                      \\
                       & &GRIT            & 939,797 (19.0\%)                     & 859,356 (17.4\%)            
    \\

\hline
\end{tabular}%
}
\vspace{-.1in}
\label{table:FN_statistics_many_discriminator}
\end{table}

\section{Additional analyses on false negative ratio (FNR) }
\begin{figure}[h!]
   \centering
   \includegraphics[width=\linewidth]{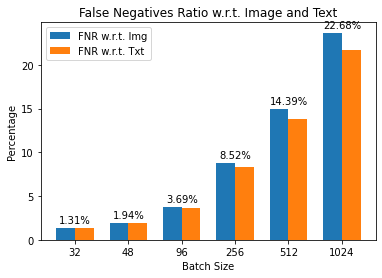}
   \vspace{-.3in}
    \caption{ FNR on 4M-Clean dataset in random sampling scenario. 
    Each annotated number is a mean of two ratios, w.r.t. image and text. }
   \label{fig:FNR_batchsize}
   \vspace{-.2in}
\end{figure}

We quantify the FNR \textit{w.r.t.} batch size under a \textit{random} sampling scenario. As shown in Figure \ref{fig:FNR_batchsize}, increasing the batch size on 4M-Clean apparently results in a higher FNR. This finding suggests the presence of a significant number of false negatives even in a \textit{random} sampling scenario with a large batch size, which is a common practice in recent VLP training schemes such as BLIP-2. Consequently, this result also underscores the applicability of MAFA in settings where hard negative sampling like GRIT is not employed.

We also measure the FNR using a considerably larger, web-crawled CC12M-Clean dataset with a default batch size of 96. In the GRIT sampling scenario, the FNR \textit{w.r.t.} images and text is 22.2\% and 20.6\%, respectively, remaining notably high. This result can serve as a proxy for the FNR on larger-scale datasets (\textit{e.g.,} LAION400M \cite{(Laion400m)schuhmann2021laion}) since they share a similar dataset construction pipeline: \textit{i.e.,} randomly sourced from the web. Therefore, these findings highlight the significance of addressing false negatives in recent large-scale models and the broad applicability of MAFA.

\section{Detailed explanation on Table \ref{table:sum_of_soft_labels} (manuscript)}
In Table \ref{table:sum_of_soft_labels} (manuscript), we present the shape of soft labels of three soft labeling methods: momentum distillation, consistency loss, and S-ITC. In this section, we explain the detailed process of constructing this table. The values in the table are obtained in the last epoch of the training with the following procedure.
\begin{enumerate}
  \item For each anchor image $i$ (resp. text $t$), a soft label is given to each text $t_k$ (resp. image $i_k$) in a batch. For the batch-size $B$, $k$ is from $1$ to $B$. For momentum distillation with queue size $Q$, $k$ is from $1$ to $B+Q$. The set of these soft labels can be represented as a $B$ (or $B+Q$)-dimensional vector, which is denoted as $\bm{y}_{a}$ for anchor $a$.
  \item $B$ (or $B+Q$) values of $\bm{y}_{a}$ are sorted in descending order, and it is denoted as $\bm{y}^\text{sorted}_{a}$.
  \item Given that the model sees $A$ anchors during an epoch of training, all $\bm{y}^\text{sorted}_{a}$s are averaged, and it is denoted as $\bm{{\bar y^\text{sorted}}}=\frac{1}{A}\Sigma_{a}\bm{y}^\text{sorted}_{a}$.
  \item Now, $B$ (or $B+Q$) values in $\bm{{\bar y^\text{sorted}}}$ are categorized into three groups: Top$1\sim 5$, Top$6\sim B$, and Top$B+1\sim Q$.
  \item The sum of the values within each category is written in the table.
\end{enumerate}

\section{Pseudocode of MAFA}
We provide pseudocode in Algorithm \ref{Appendix:code} for a thorough understanding of the MAFA framework. Although the pseudocode mostly follows the notation of the manuscript, there are some modified notations to explain the process in more detail. Specifically, $\Tilde{y}^\text{I2T-ITC}$ and $\Tilde{y}^\text{T2I-ITC}$ are used instead of $\Tilde{\bm{y}}^\text{ITC}$. Also, in the manuscript, $\Tilde{\bm{y}}^\text{ITM}$ is a $2$-dim one-hot vector for an image-text sample. If the sample is treated as a positive in ITM loss, $\Tilde{\bm{y}}^\text{ITM}$ is equal to $(1,0)$. If not, it is equal to $(0,1)$. On the other hand, in the pseudocode, $y^\text{ITM}_\text{pos}$ and $\Tilde{y}^\text{I2T-ITM}_\text{neg}$ are $B$-dim vectors in which the $b$-th element represents whether the corresponding sample is treated as positive ($1$) or not ($0$) in ITM loss.
\newcommand\mycommfont[1]{\footnotesize\ttfamily\textcolor{blue}{#1}}
\SetCommentSty{mycommfont}
\SetKwComment{Comment}{/* }{ */}

\begin{algorithm*}[hbt!]
\SetAlgoLined
\SetNoFillComment
\caption{MAFA} 
\label{Appendix:code}
 \SetKwInOut{Input}{Inputs}
\SetKwInOut{Output}{Output}
\DontPrintSemicolon
\Input{Image-text paired dataset \textit{D}, Initialized model $f_0$}
\Output{Trained model $f$}
$f \gets f_0$ \;
 \textit{Con-D} $\gets $ Train-Discriminator($D, f_0$)\;
    \For {$epoch=1,2, ..., E$}{
    \tcc{1.Similar samples are grouped by GRIT (Line 4)}
      $\Tilde{D} \gets$ GRIT($D$, $f$) \;
    \ForEach{$\{(I_b,T_b)\}_{b=1}^B \in \Tilde{D}$}{
        $\{i_b\}_{b=1}^B, \{t_b\}_{b=1}^B \gets f^\text{image}_{\text{encoder}}(\{I_b\}_{b=1}^B), f^\text{text}_{\text{encoder}}(\{T_b\}_{b=1}^B$)\;
          $\Tilde{y}^\text{I2T-ITC},\Tilde{y}^\text{T2I-ITC} \gets \mathbf{Id}_B, \mathbf{Id}_B$ \tcp*[l]{$\mathbf{Id}_B$:$B\times B$ Identity matrix }
              ${y}^\text{ITM}_\text{pos}, \Tilde{y}^\text{I2T-ITM}_\text{neg},  \Tilde{y}^\text{T2I-ITM}_\text{neg}\gets \mathbf{1}_B, \mathbf{0}_B, \mathbf{0}_B$ \tcp*[l]{$\mathbf{1}_B$:$B$-dim one vector, $\mathbf{0}_B$:$B$-dim zero vector}
              $D_\text{ITM}^\text{pos}, D_\text{ITM}^\text{neg}, D_\text{MLM}\gets\{(i_b, t_b)\}_{b=1}^B, \{\}, \{(i_b, T_b)\}_{b=1}^B$\;
            \For {$b=1,...,B$}{
                \tcc{2.A hard negative is picked per anchor (Line 11)}
                  $k \gets \arg\max_{j\ne b} {s(i_b, t_j)}$\;
                  \tcc{3.A new connection is established (Line 12-20)}

                  $p_{\text{con}} \gets \textit{Con-D}(i_b, t_{k})$\;

                \uIf {$p_\text{con} > 0.8$}{
                      $\Tilde{y}_b^{\text{I2T-ITC}}[k] \gets 1$  \tcp*[l]{$\Tilde{y}_b^{\text{I2T-ITC}}$:$b$-th row of $\Tilde{y}^{\text{I2T-ITC}}$}
                      $\Tilde{y}^{\text{I2T-ITM}}_\text{neg}[b] \gets 1$
                      \;$D_\text{MLM} \gets D_\text{MLM}\cup \{(i_b, T_k)\}$}
                \uElseIf {$0.5 < p_\text{con} < 0.8$}{
                      $k \gets \arg\max_{j\ne b,k} {\text{sim}_f(i_b, t_j)}$}
                \textbf{end}\;
                   $D_\text{ITM}^\text{neg} \gets D_\text{ITM}^\text{neg} \cup \{(i_b, t_k)\}$\;
                  $\Tilde{y}_b^{\text{I2T-ITC}} \gets \frac{\Tilde{y}_b^{\text{I2T-ITC}}}{\Sigma_j \Tilde{y}_b^{\text{I2T-ITC}}[j]}$ \;
             \tcc{(Line 11-22) are for an image anchor $i_b$. Similarly, execute for a text anchor $t_b$}

            }
            $L_{\text{S-ITC}}^\textit{ECM} \gets$ S-ITC-Loss($f, {\Tilde{y}}^\text{I2T-ITC},{\Tilde{y}}^\text{T2I-ITC}, \{(i_b, t_b)\}_{b=1}^B$) \;
              $L_{\text{ITM}}^\textit{ECM} \gets $ ITM-Loss($f, {y}^\text{ITM}_\text{pos}, \Tilde{y}^\text{I2T-ITM}_\text{neg}, \Tilde{y}^\text{T2I-ITM}_\text{neg}, D_\text{ITM}^\text{pos}\cup D_\text{ITM}^\text{neg})$\;
             $L_{\text{MLM}}^\textit{ECM} \gets $ MLM-Loss($f, D_\text{MLM}$)\;

              $f \gets$ Backward-Update($f$, $L_{\text{S-ITC}}^\textit{ECM} + L_{\text{MLM}}^\textit{ECM} + L_{\text{ITM}}^\textit{ECM})$

	    }
     }
    \KwRet{$f$} 
\end{algorithm*}

\begin{table*}[]
\caption{Comparison with state-of-the-art: fine-tuned results of IRTR on Flickr30K and MSCOCO datasets}
\vspace{-.1in}
\resizebox{\linewidth}{!}{
\begin{tabular}{ccc|ccccccccccccc}
\hline
\multirow{2}{*}{Method} &  & \multirow{2}{*}{\begin{tabular}[c]{@{}c@{}}Pre-train\\ \# Images\end{tabular}} & \multicolumn{6}{c}{MSCOCO (5K test set)}        &  & \multicolumn{6}{c}{Flickr30K (1K test set)}     \\
                        &  &                                                                                & \multicolumn{3}{c}{TR} & \multicolumn{3}{c}{TR} &  & \multicolumn{3}{c}{TR} & \multicolumn{3}{c}{IR} \\ 
                        &  &                                                                                & R@1    & R@5   & R@10  & R@1    & R@5   & R@10  &  & R@1    & R@5    & R@10 & R@1    & R@5    & R@10 \\ \hline
UNITER                  &  & 4M                                                                             & 65.7   & 88.6  & 93.8  & 52.9   & 79.9  & 88.0  &  & 87.3   & 98.0   & 99.2 & 75.6   & 94.1   & 96.8 \\
VILLA                   &  & 4M                                                                             & -      & -     & -     & -      & -     & -     &  & 87.9   & 79.30  &      & 76.3   & 73.67  &      \\
OSCAR                   &  & 4M                                                                             & 70.0   & 91.1  & 95.5  & 54.0   & 80.8  & 88.5  &  & -      & -      & -    & -      & -      & -    \\
ALBEF                   &  & 4M                                                                             & 73.1   & 91.4  & 96.0  & 56.8   & 81.5  & 89.2  &  & 94.3   & 99.4   & 99.8 & 82.8   & 96.7   & 98.4 \\
TCL                     &  & 4M                                                                             & 75.6   & 92.8  & 96.7  & 59.0   & 83.2  & 89.9  &  & 94.9   & 99.5   & 99.8 & 84     & \textbf{96.7}   & \textbf{98.5} \\
BLIP* (4M-Clean)                   &  & 4M                                                                             & 76.5   & 93.2  & 96.8  & 58.9   & 83.1  & 89.6  &  & 94.3   & 99.4   & 99.9 & 82.6   & 96.2   & 98.3 \\
GRIT-VLP*               &  & 4M                                                                             & 76.6   & 93.4  & 96.9  & 59.6   & 83.3  & 89.9  &  & 95.5   & 99.6   & 99.8 & 82.9   & 96.2   & 97.9 \\ \hline
MAFA                    &  & 4M                                                                             & 78.0   & 94.1  & 97.2  & 61.2   & 84.3  & 90.3  &  & 96.1   & 99.8   & \textbf{100}  & \textbf{84.9}  & 96.5   & 98.0 \\
MAFA (4M-Clean)         &  & 4M                                                                             & \textbf{79.4}   & \textbf{94.4}  & \textbf{97.5}  & \textbf{61.6}   & \textbf{84.5}  & \textbf{90.4}  &  & \textbf{96.2}   & \textbf{99.9}   & \textbf{100}  & 84.6     & 96.4   & 98.1 \\ \hline
ALBEF                   &  & 14M                                                                            & 77.6   & 94.3  & 97.2  & 60.7   & 84.3  & 90.5  &  & 95.9   & 99.8   & 100  & 85.6   & 97.5   & 98.9 \\
BLIP                    &  & 14M                                                                            & 80.6   & 95.2  & 97.6  & 63.1   & 85.3  & 91.1  &  & 96.6   & 99.8   & 100  & 87.2   & 97.5   & 98.8
\end{tabular}
}
\label{table:supple-overall-baseline}
\end{table*}

\begin{table*}[]
\caption{Ablation study on the proposed method: fine-tuned results of IRTR on Flickr30K and MSCOCO datasets. }
\vspace{-.1in}
\centering
\resizebox{\linewidth}{!}{%
\begin{tabular}{c|c|c|llllll|cllllll} 
\hline
\multirow{3}{*}{\begin{tabular}[c]{@{}c@{}}Pre-train\\ dataset\end{tabular}} & \multicolumn{2}{c|}{\multirow{2}{*}{MAFA}}    & \multicolumn{6}{c|}{MSCOCO (5K test set)}                                                                 & \multicolumn{1}{l}{} & \multicolumn{6}{c}{Flickr30K (1K test set)}                                                                \\
                                                                             & \multicolumn{2}{c|}{}                         & \multicolumn{3}{c}{TR}                  & \multicolumn{3}{c|}{IR}                         & \multicolumn{1}{l}{}
                                                                             & \multicolumn{3}{c}{TR}                 & \multicolumn{3}{c}{IR}                         \\ 
\cline{2-3}
                                                                             & S-ITC                 & ECM             & \multicolumn{1}{c}{R@1} & R@5   & R@10  & \multicolumn{1}{c}{R@1} & R@5   & R@10  &                      & \multicolumn{1}{c}{R@1} & R@5  & R@10  & \multicolumn{1}{c}{R@1} & R@5   & R@10      \\ 
\hline
\multirow{4}{*}{4M-Noisy}                                                    & \xmark & \xmark & 76.6                   & 93.4 & 96.9 & 59.6                   & 83.3 & 89.9 &                        & 95.5                    & 99.6 & 99.8  & 82.9                    & 96.2  & 97.9     \\
                                                                             & \xmark & \cmark & 77.4                   & 93.9 & 96.9 & 60.2                  & 83.8 & 90.4 &                        & 95.5                    & 99.5 & 99.8 & 83.7                    & 96.3  & 98.0     \\
                                                                             & \cmark & \xmark & 77.5                   & 94.3 & \textbf{97.2} & 60.5                   & 83.7 & 90.2 &                        & 96.1                    & 99.8 & 99.9  & 84.2                    & 96.3  & \textbf{98.1}     \\
                                                                             & \cmark & \cmark & \textbf{78.0}          & \textbf{94.1} & \textbf{97.2} & \textbf{61.2}          & \textbf{84.3} & \textbf{90.3} &                       & \textbf{96.1}           & \textbf{99.8} & \textbf{100.0} & \textbf{84.9}           & \textbf{96.5}  & 98.0     \\ 
\hline
\multirow{2}{*}{4M-Clean}                                                    & \xmark & \xmark & 77.7                   & 93.5 & 96.9 & 60.7                   & 83.3 & 90.1 &                        & 95.2                    & 99.6 & 99.9  & 84.2                    & 95.8 & 98.0    \\
                                                                             & \cmark & \cmark & \textbf{79.4}          & \textbf{94.4} & \textbf{97.5} & \textbf{61.6}          & \textbf{84.5} & \textbf{90.4} & 
                                                                                                  & \textbf{96.2}           & \textbf{99.9} & \textbf{100.0}   & \textbf{84.6}           & \textbf{96.4} & \textbf{98.1}   \\
\hline
\end{tabular}%
}
\label{tab:supple-full-analysis}
\end{table*}

\section{Details on Downstream tasks}
In the Supplementary Materials, the term ``total batch size" refers to the overall mini-batch size. Specifically, it represents the product of the ``number of GPUs" and the ``mini-batch size per GPU," which is calculated as $4 \times B$.
We primarily adhere to the implementation details of GRIT-VLP when performing fine-tuning. 
During fine-tuning, we employ randomly cropped images with a resolution of 384 × 384. 
Conversely, during the inference stage, we resize the images without cropping. 
Additionally, we apply the exact same RandAugment, optimizer selection, cosine learning rate decay, and weight decay with GRIT-VLP.
Following GRIT-VLP, we do not utilize a momentum encoder in the pre-training phase. 
Consequently, the momentum distillation (MD) technique is not employed for all downstream tasks,\\

\noindent{\textbf{[Image-Text Retrieval (IRTR)]}} IRTR aims to find the most similar image to a given text or text to a given image. 
Following GRIT-VLP, we do not use the momentum distillation for ITC, but use the queue and negatives from the momentum encoder when calculating ITC loss in the fine-tuning step. 
For model fine-tuning, we use the COCO and Flickr-30K datasets. 
Specifically, the COCO dataset, comprising 113,000 training images, 5,000 for validation, and another 5,000 for testing, is fine-tuned over 5 epochs. Conversely, the Flickr-30K dataset, with 29,000 training images, 1,000 for validation, and 1,000 for testing, undergoes a longer fine-tuning phase of 10 epochs.
For evaluation, we use a 5K COCO test set and  Flickr-1K set following previous works.
During the fine-tuning phase, we use a total batch size of 256 and an initial learning rate of 1e-5 for both datasets. 
Following ALBEF, during evaluation, we employ a two-step process. First, we retrieve the top-$k$ candidates by calculating image-text contrastive similarities only using uni-modal encoders. Then, we re-rank them with ITM scores. Here, $k$ is set to $256$ and $128$ for COCO and Flickr, respectively
\\
\noindent{\textbf{[Visual Reasoning (NLVR2)]}} NLVR2 is a classification task based on one caption and two images. Since the model architecture should be changed to get two images as an input, the fine-tuning step of NLVR2 requires an additional pre-training phase with the 4M-Noisy dataset for 1 epoch. 
For this pre-training phase, we employ a batch size of 256 and set the learning rate to $2e-5$, and the image resolution is set as ($256 \times 256$). 
After the single epoch pre-training phase, we fine-tune the model for 10 epochs while using a total batch size of $64$.
\\
\noindent{\textbf{[Visual Question Answering (VQA)]}} VQA is a task to obtain an answer given image and question pair. We perform experiments on the VQA2.0 dataset \cite{(VQAv2)goyal2017making}, which is divided into training, validation, and test sets with 83,000, 41,000, and 81,000, respectively.
Both the training and validation set are utilized for training. Following GRIT-VLP and ALBEF, we also include additional pairs from Visual Genome. Fine-tuning is conducted for $8$ epochs, employing a total batch size of $128$ and an initial learning rate of $2e-5$. 
For a fair comparison, the decoder only generates answers from $3192$ candidates. 

\begin{table}[hbt!]{
\caption{Comparison of computational costs. MD represents momentum distillation.}
\vspace{-.1in}
\centering
\resizebox{\linewidth}{!}{
\begin{tabular}{c|c|c|c|c} 
\hline
Model  & Time per epoch & Parameters   & Queue for MD & Queue for GRIT   \\ 
\hline
ALBEF  & 3h 10m         & 210M (MD: 210M)   & 65536 & - \\
\hline
BLIP & 3h 30m         & 252M (MD: 252M)  & 57600  & - \\
\hline
GRIT-VLP & 2h 30m         & 210M   & -  & 48000\\
\hline
MAFA & 3h 06m         & 210M (Con-D: 210M)  & -  & 48000 \\
\hline
\end{tabular}
}
\label{tab:appendix_computation}

}
\end{table}

\section{Additional Experimental Results}
\subsection{Experiments on diverse data scales}
\begin{figure}[h!]
   \centering
   \includegraphics[width=1.0\linewidth]{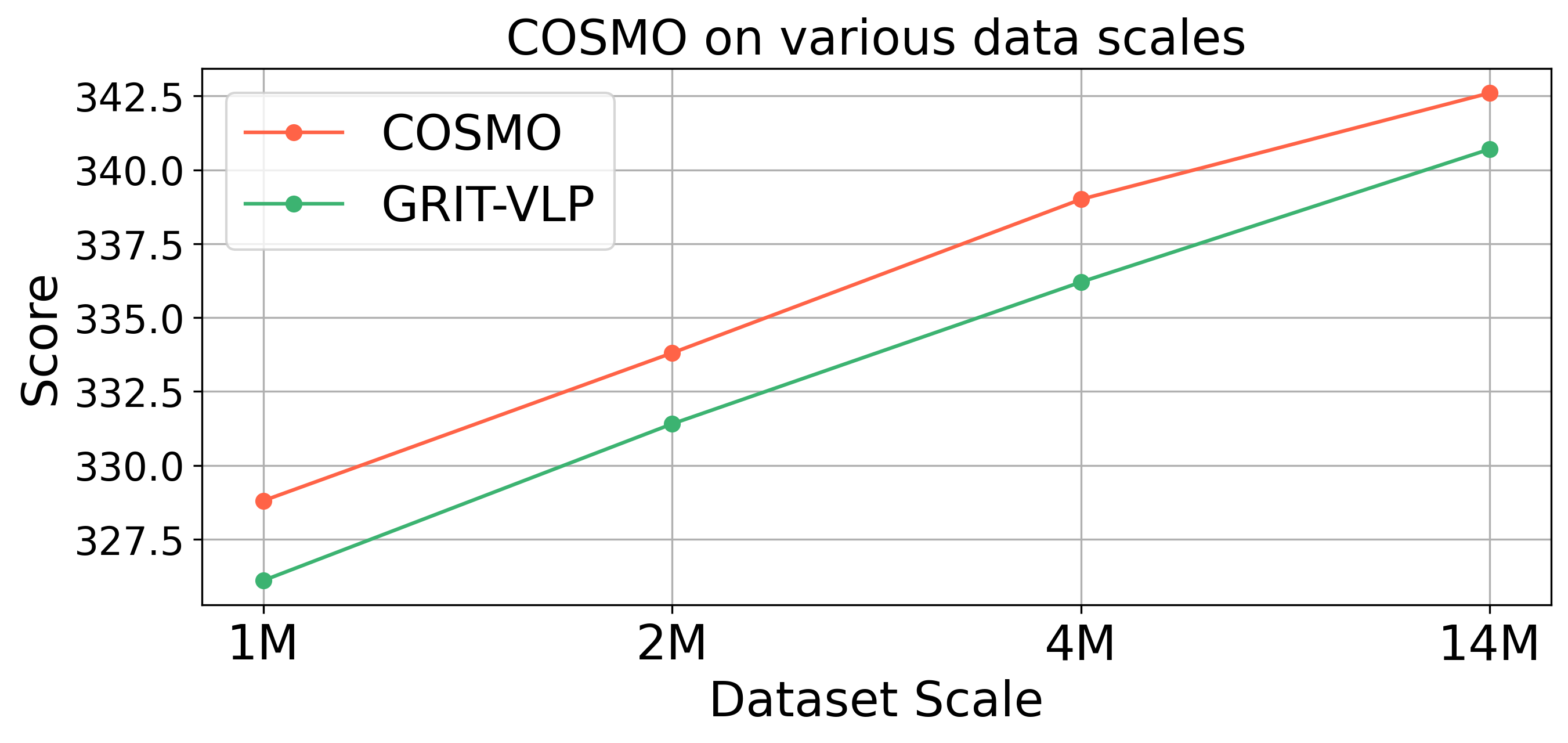}
    \caption{ The score denotes the sum of the average results of four tasks (COCO-IRTR, Flickr-IRTR, NLVR2, and VQA).}
   \label{fig:performance_comparison}
\end{figure}

We carry out additional experiments on various data scales under GRIT-sampling.
For a large-scale dataset, we utilize the 14M-Clean dataset (CC12M-Clean + 4M-Clean), and for small-scale datasets, we use 1M and 2M subsets randomly selected from the 4M-Clean dataset. To compare the overall performance, we compute the accuracy of each of the four tasks (COCO-IRTR, Flickr-IRTR, NLVR2, VQA) by averaging their respective metrics: TR/R@1, TR/R@5, TR/R@10, IR/R@1, IR/R@5, IR/R@10 for IRTR, dev and test-P for NLVR2, and text-dev and test-std for VQA. Then, we sum up the four averaged values. In Figure \ref{fig:performance_comparison}, we observe that MAFA consistently achieves considerable performance improvements across all data scales compared to GRIT-VLP, again demonstrating the robustness of MAFA \textit{w.r.t.} data scale variations. 

\subsection{Computation comparison} \label{sec:appendix_computation}
We describe the computational cost for pre-training. Table \ref{tab:appendix_computation} shows the pre-training time per epoch, the number of parameters of the model, and queue size. Here, the ``MD" denotes additional model parameters for the momentum model, and ``Con-D" denotes additional model parameters for \textit{Con-D} in the pre-training phase. As pointed out in the manuscript (Section \ref{sec:method}), although MAFA becomes relatively slower than the GRIT-VLP, it is still competitive with ALBEF, and faster than BLIP.

\subsection{Detailed experiments on overall framework}
We present a comprehensive comparison of various baselines on image-text retrieval tasks in Table \ref{table:supple-overall-baseline}. 
We observe a similar tendency in the manuscript (Section \ref{sec:experiments}).
Namely, MAFA mostly surpasses other baselines in performance when pre-trained on the 4M-Noisy dataset, outperforming even ALBEF pre-trained on the much larger dataset (14M). Moreover, in the case of the 4M-Clean dataset, MAFA again shows a significant improvement over BLIP, which is pre-trained on the same dataset.

\subsection{Analysis on search space}
When applying GRIT sampling, the degree of similarity between samples within a batch is influenced by the search space since GRIT groups together similar samples within that specific search space. 
Consequently, an increase in the search space leads to a higher number of false negatives as shown in Figure \ref{fig:motivation_plot} (manuscript). We observe that the average IRTR performance of GRIT-VLP is decreased as the search space increases, as shown in Table \ref{table:searchspace_GRIT_MAFA}. 
In contrast, MAFA shows a performance improvement with larger search spaces and consistently outperforms GRIT-VLP regardless of the search space.
We believe this result shows the effectiveness of MAFA in managing \textit{false negatives}.

\begin{table}[hbt!]
\caption{Performance comparison of GRIT-VLP and MAFA  on COCO using different search space sizes.}
\vspace{-.1in}
\resizebox{\linewidth}{!}{%
\begin{tabular}{l|c|lllllll}
\hline
\multirow{3}{*}{Method}   & \multirow{3}{*}{$M$} & \multicolumn{7}{c}{MSCOCO (5K test set)}                                                                                                                                                                              \\
                          &                               & \multicolumn{3}{c}{TR}                                                       & \multicolumn{3}{c}{IR}                                                         & \multicolumn{1}{c}{\multirow{2}{*}{Avg}} \\
                          &                               & \multicolumn{1}{c}{R@1} & \multicolumn{1}{c}{R@5} & \multicolumn{1}{c}{R@10} & \multicolumn{1}{c}{R@1} & \multicolumn{1}{c}{R@5} & \multicolumn{1}{c}{R@10}   & \multicolumn{1}{c}{}                     \\ \hline
\multirow{3}{*}{GRIT-VLP} & 960                           & 76.7                   & 93.9                   & 97.2                    & 59.6                   & 83.5                   & \multicolumn{1}{l|}{90.1} & 83.5                                    \\
                          & 4800                          & 76.6                   & 93.4                   & 96.9                    & 59.6                   & 83.3                   & \multicolumn{1}{l|}{89.9} & 83.3                                    \\
                          & 48000                         & 77.0                   & 93.1                   & 96.6                     & 59.2                   & 83.1                   & \multicolumn{1}{l|}{89.8} & 83.1                                    \\ \hline
\multirow{3}{*}{MAFA}     & 960                           & 77.6                    & \textbf{94.4}                   & 97.1                    & 60.7                  & 84.1                   & \multicolumn{1}{l|}{90.2} & 84.0                                  \\
                          & 4800                          & 78.0                   & 94.1                   & 97.2                    & 61.2                   & 84.3                   & \multicolumn{1}{l|}{90.3} & 84.2                                    \\
                          & 48000                         & \textbf{78.3}                   & 94.2                   & \textbf{97.3}                     & \textbf{61.4}                   & \textbf{84.4}                   & \multicolumn{1}{l|}{\textbf{90.5}} & \textbf{84.4}                                    \\ \hline
\end{tabular}
}
\vspace{-.1in}
\label{table:searchspace_GRIT_MAFA}
\end{table}


\subsection{Detailed experiments on ECM} \label{sec:appendix_fn_filter}
\noindent{\textbf{[Comparative analysis with ECM variants]}}
We conduct a comparative analysis of different uses of new missing positives constructed from ECM.
Namely, we assess the performance of models when new positives are exclusively incorporated in ITC, ITM, and MLM, respectively. 
As can be verified in Table \ref{tab:ECM-ablation}, the usage of new positives is effective for each of these objectives.
We believe the relatively marginal impact of new positives in ITC arises from the possible inclusion of additional false negatives in the mini-batch, which highlights the necessity of S-ITC.
Moreover, their combined usage across three objectives leads to more improvements. 
We believe this result clearly shows the benefits of using new missing positives constructed from ECM, across all objectives.
Note that all variants do not use S-ITC as objectives here. 
\begin{table}[htb!]
\caption{Comparison of COCO performance with ECM-variants}
\vspace{-.1in}
\resizebox{\columnwidth}{!}{%
\begin{tabular}{ccc|cccccc|c}
\hline
\multicolumn{3}{c|}{\multirow{2}{*}{ECM}}                 & \multicolumn{6}{c|}{MSCOCO (5K test set)}        &      \\
\multicolumn{3}{c|}{}                                     & \multicolumn{3}{c}{TR} & \multicolumn{3}{c|}{IR} &      \\ \cline{1-3}
\multicolumn{1}{c|}{ITC} & \multicolumn{1}{c|}{ITM} & MLM & R@1    & R@5    & R@10 & R@1    & R@5    & R@10  & Avg  \\ \hline
\multicolumn{1}{c|}{\xmark}   & \multicolumn{1}{c|}{\xmark}   & \xmark   & 76.6   & 93.4   & 96.9 & 59.6   & 83.3   & 89.9  & 83.3 \\
\multicolumn{1}{c|}{\cmark}   & \multicolumn{1}{c|}{\xmark}   & \xmark   & 77.1    & 93.5       & 96.8     & 59.7    &    83.4 & 90.1      & 83.4     \\
\multicolumn{1}{c|}{\xmark}   & \multicolumn{1}{c|}{\cmark}   & \xmark   & 77.1   & 93.5   & \textbf{97.3} & 59.9   & 83.7   & 90.0  & 83.6 \\
\multicolumn{1}{c|}{\xmark}   & \multicolumn{1}{c|}{\xmark}   & \cmark   & \textbf{77.4}   & 93.7   & 97.0 & 59.9   & 83.2   & 89.9  & 83.5 \\
\multicolumn{1}{c|}{\cmark}   & \multicolumn{1}{c|}{\cmark}   & \cmark   & \textbf{77.4}   & \textbf{93.9}   & 96.9 & \textbf{60.2}   & \textbf{83.8}   & \textbf{90.4}  & \textbf{83.8} \\ \hline
\end{tabular}
}
\label{tab:ECM-ablation}
\end{table}

\begin{table}[htb!]
\caption{Comparison of COCO performance using different threshold $\tau$}
\vspace{-.1in}
\resizebox{\columnwidth}{!}{%
\begin{tabular}{c|cc|cc|cc}
\hline
\multirow{2}{*}{$\tau$} & \multicolumn{2}{c|}{COCO R@1} & \multicolumn{2}{c|}{NLVR2} & \multicolumn{2}{c}{VQA} \\ \cline{2-7} 
                        & IR            & TR            & dev         & test-P       & test-dev   & test-std   \\ \hline
0.5                     & \textbf{78.0}          & \textbf{61.2}          & 82.06       & \textbf{82.35}        & \textbf{75.60}      & \textbf{75.78}      \\
0.8                    & \textbf{78.0}          & \textbf{61.2}         & \textbf{82.52}       & 82.08        & 75.55      & 75.77      \\ \hline
\end{tabular}
}
\label{tab:ECM-tau}
\end{table}

\noindent{\textbf{[Robustness of MAFA for varying threshold $\tau$]}}
We evaluate the impact of choosing different thresholds $\tau$ for \textit{Con-D}. Table \ref{tab:ECM-tau} demonstrates that ECM process exhibits stability across different thresholds. 
Here, if $\tau$ is set to $0.5$, the re-sampling strategy for ITM is omitted. 


\noindent{\textbf{[Comparison with oracle \textit{Con-D}]}}
We measure the performance of MAFA using different \textit{Con-D}: one pre-trained on the 4M-Noisy dataset and the other obtained from BLIP (pre-trained with 129M dataset). 
In Table \ref{tab:oracle-analysis}, we observe that MAFA exhibits only a little performance gap with the MAFA (Oracle). 
We believe this result shows that the \textit{Con-D} constructed with a 4M-Noisy dataset is sufficient to reliably identify \textit{false negatives} within that particular dataset.
This aligns with the findings of BLIP, where it was shown that a filter trained with a noisy dataset of the same scale can effectively handle \textit{false positives} within that specific dataset. 

\begin{table}[h!]
\caption{Comparison with MAFA (Oracle) which uses the strong \textit{Con-D} pre-trained with 129M data from BLIP}
\vspace{-.1in}
\resizebox{\columnwidth}{!}{%
\begin{tabular}{c|c|cc|cc}
\hline
\multirow{2}{*}{Dataset}  & \multirow{2}{*}{Method} & \multicolumn{2}{c|}{COCO R@1} & \multicolumn{2}{c}{Flickr R@1} \\
                          &                         & TR            & IR            & TR            & IR             \\ \hline
\multirow{3}{*}{4M-Noisy} & GRIT                    & 76.6         & 59.6         & 95.5              & 82.9              \\
                          & MAFA                    & 78.0         & \textbf{61.2}         & \textbf{96.1}          & \textbf{84.9}             \\
                          & MAFA (Oracle)           & \textbf{78.5 }        & \textbf{61.2}        &    95.8     &   84.2       \\ \hline
\end{tabular}
}
\label{tab:oracle-analysis}
\end{table}

\subsection{Additional analyses on ITC} 
\noindent\textbf{{[Effect of Queue in Momentum Distillation]}} \label{sec:appendix:UITC} In this section, we present additional experiments and analyses that delve deeper into the topics discussed in the manuscript. Given the significance of larger batch sizes in contrastive learning, it is standard practice to maintain a sufficient number of negative samples by employing a queue. To investigate the impact of the queue, we compare the performance of momentum distillation with a queue (MD) and without a queue (MD\textsubscript{NQ}).

Table \ref{table:ITC-supple} shows the results of our observations regarding the effect of the queue in the GRIT sampling scenario. We observe that the queue has a detrimental effect on performance in this particular setting. This can be attributed to the substantial increase in the number of negatives, which consequently leads to significantly smaller labels assigned to negatives (compared to the case without a queue).
The impact of the queue highlights the crucial requirement of non-zero soft labels in the GRIT sampling scenario. 

\begin{table}[hbt!]
\caption{Comparison of smoothing for ITC.}
\vspace{-.1in}
\centering
\resizebox{\columnwidth}{!}{%
\begin{tabular}{c|c|cc|cc}
\hline
\multirow{2}{*}{Method}   & \multirow{2}{*}{Label smoothing} & \multicolumn{2}{c|}{COCO R@1}         & \multicolumn{2}{c}{Flickr R@1}    \\
                          &                              & TR                         & IR       & TR                        & IR    \\ \hline
\multirow{4}{*}{ALBEF}    & X                            & \textbf{74.4}                       & 57.6    & \textbf{93.5}                      & 81.7 \\
                          & MD\textsubscript{NQ} & 73.8                      & \textbf{57.9} & 93.3                      & 81.5 \\
                          & MD                     & 74.2                       & 57.4 & \textbf{93.5}                      & \textbf{81.9} \\
                          & S-ITC                    & 73.5                      & 56.1 & 92.9                      & 79.9 \\ \hline
\multirow{4}{*}{GRIT-VLP} & CS             & \multicolumn{1}{c}{76.6} & 59.6 & \multicolumn{1}{c}{95.5} & 82.9  \\
                          & MD\textsubscript{NQ} & 77.1                      & 59.8 & 95.5                      & 83.8 \\
                          & MD                     & 76.1                      & 58.9 & 94.4                      & 82.7  \\
                          & S-ITC                    & \textbf{77.5}                      & \textbf{60.5} & \textbf{96.1}                      & \textbf{84.2 } \\ \hline
\end{tabular}%
}
\label{table:ITC-supple}
\end{table}

\noindent\textbf{{[Effect of Mixing parameter $\alpha$]}}
As depicted in Table \ref{table:sum_of_soft_labels} (manuscript), the failure of MD and CS to provide soft labels can be attributed to different reasons. MD fails due to the large size of the queue (48000), while CS excessively concentrates only on a few similar samples. Adjusting the parameter $\alpha$ alone does not resolve this issue, as the underlying problem lies in the inclination of the model to favor the closest few samples.
Moreover, naively increasing the value of $\alpha$ poses another challenge in training the model which amplifies the portion of uncertain labels of the momentum model (or the model itself) rather than ground-truth labels.

On the contrary, as shown in Table \ref{table:ITC_alpha}, the S-ITC method demonstrates robustness across a wide range of $\alpha$ values from 0.1 to 0.7. This highlights the distinct approach of S-ITC, which deviates from the inclination of the model towards the closest samples and effectively assigns non-zero labels to majorities of the negatives in the GRIT sampling scenario.

\begin{table}[hbt!]
\caption{Comparison of COCO performance using varying $\alpha$ of S-ITC in the GRIT sampling scenario, with a fixed training epoch of 10.}
\vspace{-.1in}
\resizebox{\linewidth}{!}{%
\begin{tabular}{c|cccccc|c}
\hline
      & \multicolumn{6}{c|}{MSCOCO (5K test set)}            &         \\
      & \multicolumn{3}{c}{TR} & \multicolumn{3}{c|}{IR}     &         \\
$\alpha$ & R@1    & R@5   & R@10  & R@1     & R@5     & R@10    & Avg \\ \hline
0.1   & 75.5  & \textbf{93.3} & \textbf{96.9} & 58.4 & 82.7 & 89.3 & \textbf{82.7} \\
0.3   & \textbf{75.8}  & 93.0 & 96.4 & 58.8 & 82.7 & 89.4 & \textbf{82.7} \\
0.5   & 75.6  & 92.9 & 96.6  & \textbf{59.0} & \textbf{82.8 }& \textbf{89.6} & \textbf{82.7} \\
0.7   & 75.3  & 93.1  & 96.7 & 58.3    & 82.6   & 89.3   & 82.6  \\
0.9   & 73.0  & 92.1  & 96.2 & 56.4   & 81.3   & 88.6   & 81.3 \\ \hline
\end{tabular}
}
\label{table:ITC_alpha}
\end{table}

\begin{table*}[t!]
\caption{Results on NoCaps and COCO Caption. All methods are fine-tuned. C: CIDEr, S: SPICE, B@4: BLEU@4.
}
\vspace{-.1in}
\resizebox{\linewidth}{!}{%
\begin{tabular}{c|cccccccc|cc}
\hline
\multirow{3}{*}{Model} & \multicolumn{8}{c|}{NoCaps Zero-shot (validation set)}                                                                          & \multicolumn{2}{c}{\multirow{2}{*}{\begin{tabular}[c]{@{}c@{}}COCO Fine-tuned\\ Karpathy test\end{tabular}}} \\
                       & \multicolumn{2}{c}{in-domain} & \multicolumn{2}{c}{near-domain} & \multicolumn{2}{c}{out-domain} & \multicolumn{2}{c|}{overall} & \multicolumn{2}{c}{}                                                                                        \\
                       & C              & S            & C              & S              & C              & S             & C             & S            & B@4                                                   & C                                                   \\ \hline
BLIP-2                  & 109.55               & 14.46             & 105.83               & 14.26               & 106.33               & 13.51              & 106.47              & 14.14             & 40.7                                                      & 139.7                                                    \\
BLIP-2+MAFA            & \textbf{114.67}               & \textbf{15.19}             & \textbf{111.74}               & \textbf{14.88}               & \textbf{113.29}               & \textbf{14.27}              & \textbf{112.48}              & \textbf{14.80}             &  \textbf{41.6}                                                 & \textbf{142.5}                                                    \\ \hline
\end{tabular}
}
\label{tab:BLIP-2-captioning}
\end{table*}
\subsection{Details on experiments with BLIP-2} \label{sec:appendix_BLIP-2}
BLIP-2 aims to propose an efficient vision-language pre-training framework that connects off-the-shelf frozen pre-trained image encoders and frozen large language models.
To bridge the modality gap, BLIP-2 adopts a lightweight Querying Transformer (Q-Former), which is pre-trained in two stages.
In \textit{stage 1}, Q-Former is pre-trained to extract visual features relevant to text from a frozen image encoder with ITC, ITM, and (autoregressive) LM objectives. Thus, the training objectives are almost the same as those of ALBEF \cite{(ALBEF)li2021align} and BLIP \cite{(BLIP)li2022blip}.
In \textit{stage-2}, Q-Former is connected to a frozen LLM and pre-trained with LM loss to generate the text conditioned on the visual representation from Q-Former.

We mainly follow the default implementation setting of BLIP-2. Namely, we use ViT-G/14 from EVA-CLIP \cite{(EVA)fang2023eva} as the vision encoder and OPT-2.7B \cite{(OPT)zhang2022opt} as a language decoder. 
Moreover, Q-Former is initialized with BERT-base \cite{(BERT)devlin2018bert} and has 32 learnable queries per single representation. 
For training and evaluation of BLIP-2, we use 8 A100 GPUs. For \textit{stage 1}, we use the total batch size as $384$ which is the same as the previous setting (in the manuscript). 
For \textit{stage 2}, we use the total batch size as $512$. For image-text retrieval fine-tuning, we use $112$ as the total batch size. 
For other hyper-parameters, we use exactly the same as the default setting from BLIP-2 \cite{(BLIP-2)li2023blip}.
For GRIT sampling, we choose $3840$ as search space, which is $10$ times of the batch size. 
Since BLIP-2 uses 32 queries per image, we extract a single query representation that has a maximum similarity with its corresponding text and then use this representation to find similar examples for GRIT sampling.

For MAFA, to accelerate the efficiency of the experiment, we omit the re-sampling strategy in ITM and usage of additional positives, and we set $\alpha$ as $0.2$ for S-ITC.
Note that GRIT sampling and MAFA are only applied in \textit{stage-1}.
In both two stages, we use the 4M-Noisy dataset. 
Moreover, as mentioned in Section \ref{sec:experiments} (manuscript), since the exclusive use of GRIT sampling causes the failure of learning, we omit the results of ``BLIP-2+GRIT'' in \textit{stage 2}. In addition, to adopt MAFA, we utilize the original BLIP-2 model for \textit{Con-D}, which is pre-trained with the 4M-Noisy dataset and then fine-tuned with the COCO dataset. 
For zero-shot VQA tasks, following BLIP-2, we utilize the prompt ``Question: Answer:'' and beam search with beam width 5 and set length-penalty to -1 for all models.

\subsection{Additional results with BLIP-2} \label{sec:appendix_BLIP-2}
\noindent{\textbf{[Results on fine-tuned image captioning]}}
We fine-tune models on COCO with $5$ epochs with a total batch size of $128$.
We use the prompt ``a photo of'' as the initial input for the LLM decoder (OPT-2.7B model) and train with autoregressive LM loss. 
For all other hyper-parameters, we use the exact same hyper-parameters as BLIP-2. 
In fine-tuning, the parameters of the Q-Former and image encoder are only updated while those of LLM are kept frozen.
We evaluate models on both the Karpathy test split of MSCOCO and zero-shot transfer ability to NoCaps dataset \cite{(nocaps)agrawal2019nocaps}.
The results, which can be observed in Table \ref{tab:BLIP-2-captioning}, indicate a similar trend in zero-shot ability. 
Namely, MAFA significantly enhances the captioning ability. 
Note that MAFA is exclusively used in \textit{stage 1} and not used in \textit{stage 2} and the fine-tuning stage.

\begin{table}[]
\caption{Effectiveness of extra positives from ECM (\textit{stage 2}). 
}
\vspace{-.1in}
\resizebox{\columnwidth}{!}{%
\begin{tabular}{c|c|c|c|cc|c}
\hline
\multirow{2}{*}{Model} & VQAv2 & OK-VQA & GQA      & \multicolumn{2}{c|}{\begin{tabular}[c]{@{}c@{}}COCO zero-shot\\ Karpathy test\end{tabular}} &\multirow{2}{*}{Sum} \\
                       & val   & test   & test-dev & BLEU@4                                       & CIDEr         &                          \\ \hline
4M                  & 46.6  & 23.8   & 29.1     & 35.6                                        & 118.8             &  253.9                         \\ \hline
repeated-6M          & \textbf{47.9}  & 24.9   & 30.8     & 35.8                                        & 118.3   & 257.7
\\ \hline
ECM-6M          & 47.4  & \textbf{27.0}   & \textbf{30.9}     & \textbf{37.5}                                        & \textbf{123.6}   & \textbf{266.4}
\\ \hline

\end{tabular}
}
\label{tab:BLIP-2-extra}
\end{table}

\noindent{\textbf{[Results of \textit{stage-2} with extra positives from ECM]}} To further investigate the wider applicability of ECM in VLP models (w/o ITC and ITM), we conduct additional experiments on BLIP-2 \textit{stage-2} model with extra positives generated by ECM. Here, in contrast to the results in Tables \ref{table:with_BLIP-2}, \ref{tab:BLIP-2-zero} and \ref{tab:BLIP-2-captioning} (manuscript), \textit{stage-1} model is trained without MAFA. Namely, before training \textit{stage-2 model}, by applying GRIT-sampling and using the frozen BLIP-2 stage-1 model as our \textit{Con-D}, we generate 2M additional positives with a single forward pass, augmenting the original 4M-Noisy dataset. As reported in Table \ref{tab:BLIP-2-extra},  we observe the \textit{stage-2} model, which is trained on this ``ECM-6M'' (4M-Noisy + ECM-generated 2M) dataset, significantly outperforms the baseline trained on ``repeated-6M'' (4MNoisy + 2M sampled from the same 4M-Noisy) dataset. The scores are 266.4 vs. 257.7 in which the zero-shot results across VQA, OK-VQA, GQA, and COCO captioning tasks are summed. We believe this result validates the effectiveness of ECM-generated positives and underscores the general applicability of our framework.

\section{Additional examples of new positive connections by ECM in training}
\begin{figure}[h!]
   \centering
   \includegraphics[width=0.8\linewidth]{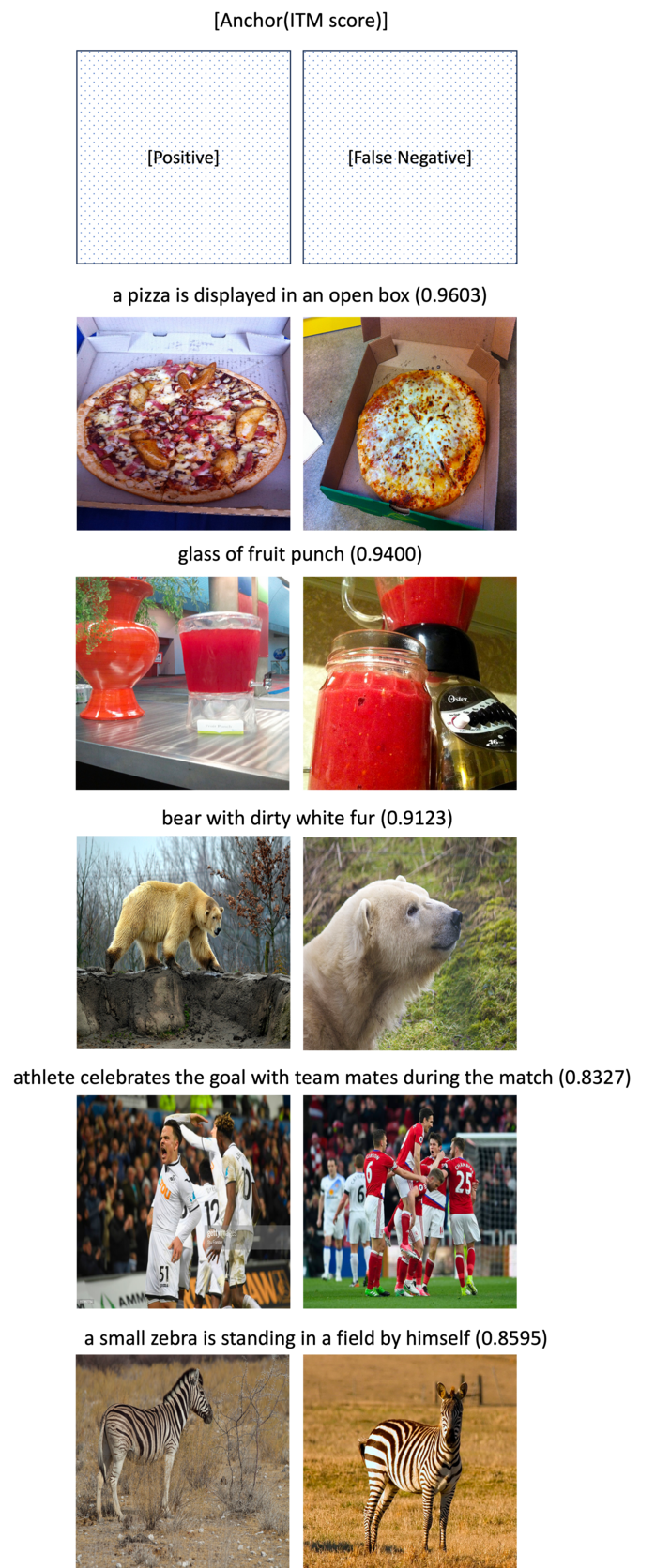}
    \caption{Examples of new positive connections constructed by ECM with respect to the texts during training.}

   \label{fig:FN_wrttxt}
\end{figure}

\begin{figure}[h!]
   \centering
   \includegraphics[width=0.9\linewidth]{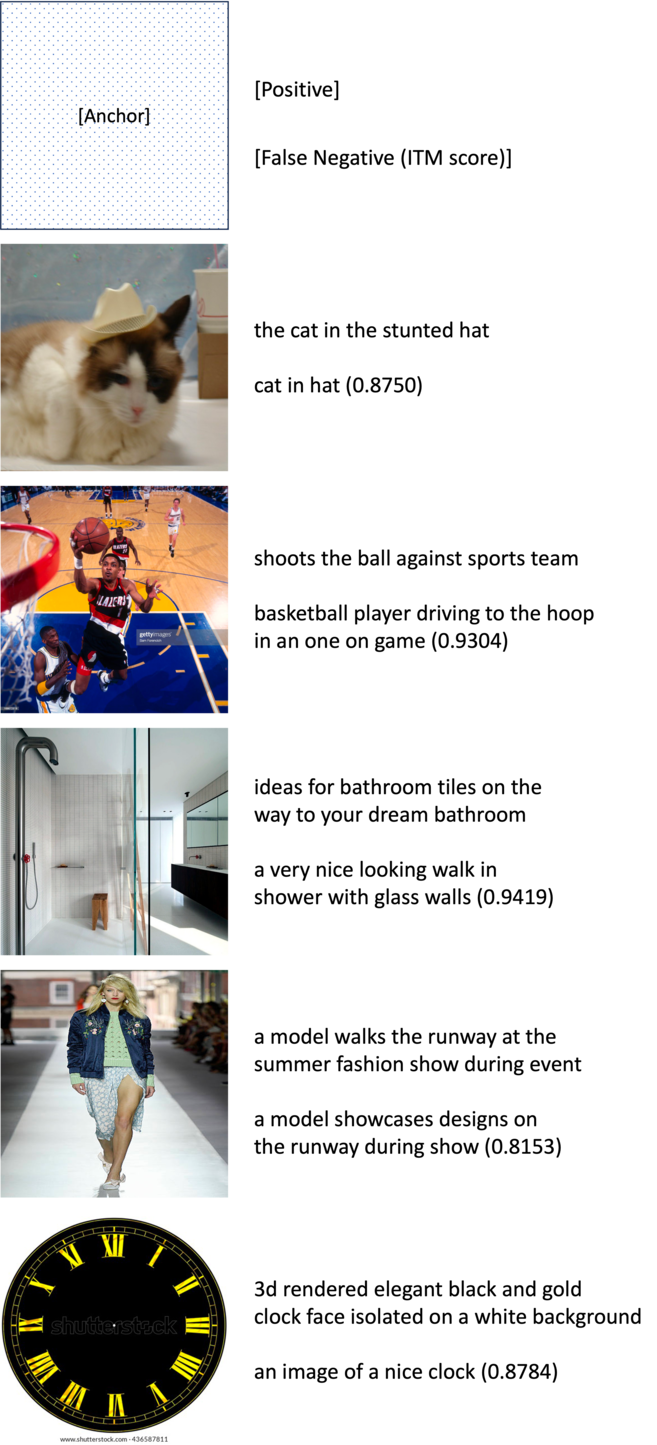}
        \caption{Examples of new positive connections constructed by ECM with respect to the images during training.}

   \label{fig:FN_wrtimg}
\end{figure}

We provide additional examples of new positive connections by ECM during training. Figures \ref{fig:FN_wrttxt} and \ref{fig:FN_wrtimg} show anchors ([Anchor]), their corresponding pairs ([Positive]), and new positives ([False Negative]) constructed by ECM during training. The number in parentheses indicates the ITM score between the anchor and the false negative computed by \textit{Con-D}.

\newpage